\documentclass[11pt]{article}

\usepackage{fullpage}
\usepackage{CJKutf8} 
\usepackage[utf8]{inputenc}
\usepackage{setspace}
\usepackage{parskip}
\usepackage{titlesec}
\usepackage[section]{placeins}
\usepackage{xcolor}
\usepackage{breakcites}
\usepackage{lineno}
\usepackage{hyphenat}

\usepackage{times}

\PassOptionsToPackage{hyphens}{url}
\usepackage[colorlinks = true,
            linkcolor = blue,
            urlcolor  = blue,
            citecolor = blue,
            anchorcolor = blue]{hyperref}
\usepackage{etoolbox}

\usepackage{natbib}

\titlespacing{\section}{0pt}{*3}{*1}
\titlespacing{\subsection}{0pt}{*2}{*0.5}
\titlespacing{\subsubsection}{0pt}{*1.5}{0pt}

\usepackage{graphicx}
\usepackage[space]{grffile}
\usepackage{latexsym}
\usepackage{textcomp}
\usepackage{longtable}
\usepackage{tabulary}
\usepackage{booktabs,array,multirow}
\usepackage{amsfonts,amsmath,amssymb}
\providecommand\citet{\cite}
\providecommand\citep{\cite}

\newif\iflatexml\latexmlfalse

\AtBeginDocument{\DeclareGraphicsExtensions{.pdf,.PDF,.eps,.EPS,.png,.PNG,.tif,.TIF,.jpg,.JPG,.jpeg,.JPEG}}

\usepackage[utf8]{inputenc}
\usepackage[english]{babel}

\begin{document}
\begin{CJK}{UTF8}{gbsn}

%\title{Generative Models of Neural Noise Reveal the Structure of Higher-Order Representations}

%Generative denoising models to reveal higher-order representations via noninvasive neuroimaging

\title{Characterizing higher-order representations through generative diffusion models explains human decoded neurofeedback performance}

%Generative diffusion models to explain human decoded neurofeedback performance via characterizing higher-order this is too long 

%generative diffusion models capture individual differences in higher-order representations

\vspace{-1em}

  \date{}

\begingroup
\let\center\flushleft
\let\endcenter\endflushleft
\maketitle
\endgroup

% \linenumbers

\doublespacing

\sloppy

\textbf{Abbreviated title (50 character max)}: Generative Models of Higher-Order Representations

\textbf{Author Names and Affiliations:}~ 
Hojjat Azimi Asrari\(^{1}\), 
Megan A. K. Peters\(^{1,2,3,4,5,6}\)

\(^{1}\): Department of Cognitive Sciences, University of California Irvine, Irvine, CA 92617, USA

\(^{2}\): Department of Logic \& Philosophy of Science, University of California Irvine, Irvine, CA 92617, USA

\(^{3}\): Center for the Neurobiology of Learning \& Memory, University of California Irvine, Irvine, CA 92617, USA

\(^{4}\): Center for Theoretical Behavioral Sciences, University of California Irvine, Irvine, CA 92617, USA

\(^{5}\): Department of Experimental Psychology, University College London, London WC1H 0AP, England

\(^{6}\): Program in Brain, Mind, \& Consciousness, Canadian Institute for Advanced Research, Toronto, ON M5G 1M1, Canada

\textbf{Number of pages:} 25

\textbf{Number of figures:} 7

\textbf{Number of Words}

\textbf{Abstract:} 170

\textbf{Introduction:} 632

\textbf{Discussion:} 1458

\textbf{Conflict of interest statement:} M.A.K.P. is a consultant for the for-profit entity Conscium, Inc., which seeks to pioneer safe, efficient artificial intelligence and which played no role in this project's conceptualization, analyses, interpretation, or writing. The authors declare no conflicts of interest.

\textbf{Acknowledgments:} This project was supported in part by a fellowship (to M.A.K.P.) from the Canadian Institute for Advanced Research Program in Brain, Mind, \& Consciousness and a grant from the Templeton World Charity Foundation (``An adversarial collaboration to empirically evaluate higher-order theories of consciousness", TWCF 22032, to M.A.K.P.). The funding sources had no role in the design, implementation, or interpretation of the work presented here.

\newpage

\section*{Abstract} % (250 Words Maximum)}

Brains construct not only ``first-order'' representations of the environment but also ``higher-order'' representations about those representations -- including higher-order uncertainty estimates that guide learning and adaptive behavior. Higher-order \textit{expectations} about representational uncertainty -- i.e., learned through experience -- may play a key role in guiding behavior and learning, but their characterization remains empirically and theoretically challenging. Here, we introduce the Noise Estimation through Reinforcement-based Diffusion (NERD) model, a novel computational framework that trains denoising diffusion models via reinforcement learning to infer distributions of noise in functional MRI data from a decoded neurofeedback task, where healthy human participants learn to achieve target neural states. We hypothesize that participants accomplish this task by learning about and then minimizing their own representational uncertainty. We test this hypothesis with NERD, which mirrors brain-like unsupervised learning. Our results show that NERD outperforms backpropagation-trained control models in capturing human performance.
%($R^{2} = 0.782$ vs.\ $0.321$)
%with explanatory power enhanced by clustering learned noise distributions.
%($R^{2} = 0.869$ vs.\ $0.582$). 
Importantly, our results also reveal individual differences in expected-uncertainty representations that predict task success, demonstrating NERD's utility as a powerful tool for probing higher-order neural representations.

\section*{Keywords}
decoded neurofeedback; denoising diffusion models; reinforcement learning; fMRI; higher-order representations

\section*{Significance Statement } % (120 Words Maximum)

{\label{866667}}

The human brain constantly estimates not only external world states, but also the uncertainty in these estimations. However, understanding how the brain represents its own uncertainty remains a challenging scientific question. We propose a novel generative artificial intelligence framework, the Noise Estimation through Reinforcement-based Diffusion (NERD) model, which simulates how the brain learns about its own noise in an unsupervised manner. By applying NERD to human brain imaging data, we demonstrate that it successfully captures how subjects learn to self-regulate their brain activity through minimizing uncertainty. Furthermore, NERD achieves better performance than control models trained with traditional backpropagation. Our work provides a powerful new tool to reveal how the mind learns to represent the world and itself.

\section*{Introduction} % 650 words maximum

{\label{872912}}

Neural representations encode information about our environments or internal states, guiding perception, decision-making, and behavior \citep{baker2022three, tarr2002visual, shahdoust2024interictal, libowitz2025increased}. ``First-order" representations encode external environmental properties, whereas ``higher-order" representations encode properties of these first-order representations, such as their strength, source, or uncertainty \citep{baker2022three, tarr2002visual, squire1991medial, miller2001integrative, cleeremans2007conscious, Brown2019Understanding, fleming2020awareness, lau2019consciousness, michelperceptual}. Higher-order representations of uncertainty are particularly critical for metacognition and learning, for example influencing how internal world models are updated \citep{Meyniel2015Sense, meyniel2015confidence, Froemer2021Response, Hainguerlot2018Metacognitive,Meyniel2017Brain, guggenmos2016mesolimbic, Guggenmos2022Reverse}. However, studying them is challenging due to their indirect relationship with observable variables and a lack of consensus regarding the computations that give rise to them \citep{peters2025introspective, shekhar2024humans, Boundy-Singer2023Confidence, mamassian2018confidence, mamassian2022modeling, Mamassian2024CNCB, peters2017perceptual,petersAzimiTheory}.

A recent proposal suggests that higher-order estimates of first-order uncertainty incorporate both current uncertainty levels and more general expectations about uncertainty that have been learned through experience \citep{petersAzimiTheory,winter2022variance}. This proposal unifies metacognitive inference with the widely-successful Bayesian framework for describing and explaining perception and cue combination \citep{Knill1996}. Despite this, representations of \textit{expected} uncertainty remain largely unexplored due to their relative inaccessibility. While neural correlates of behaviorally-reported uncertainty estimates exist, they reflect the outputs of -- rather than the inputs to -- metacognitive computations \citep{Bang2018Distinct,cortese2017decoded,kianishadlen2009,odegaard2018pnas,peters2017perceptual,walker2023studying, peters2022towards}. Furthermore, existing methods that directly quantify uncertainty in first-order representations, for example through neural decoding approaches \citep{ma2006bayesian, ma2009population,vanbergen2021tafkap, Kay2024Disentangling,prince2022improving}, fail to capture how the brain estimates its own representational uncertainty.%, which would be necessary for characterizing these higher-order uncertainty estimation processes.

\begin{figure*}[ht]
    \centering
    {\includegraphics[width=\textwidth]{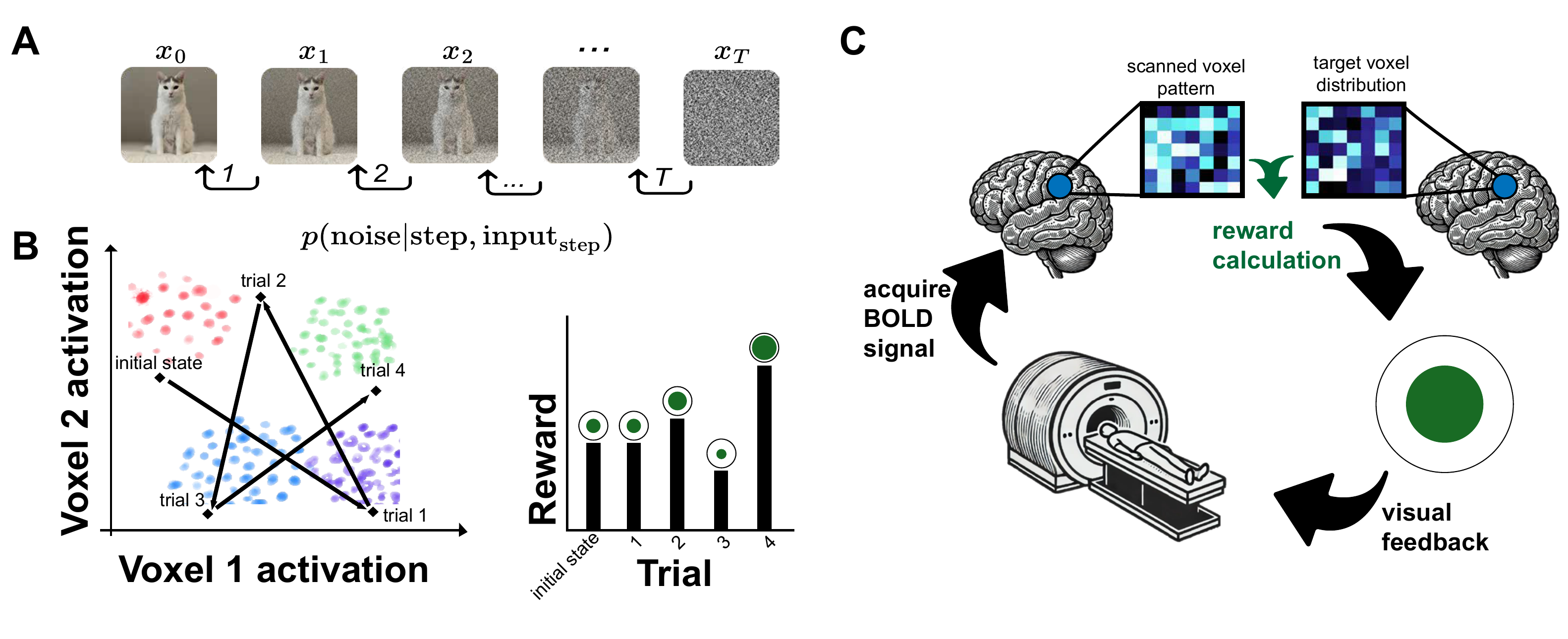}}
\caption{\textbf{Cartoons illustrating the denoising process learned by diffusion models and the closed-loop real-time neurofeedback training procedure.} 
(A) Denoising diffusion models are trained to learn the distribution of pixel noise conditioned on the current denoising step $t$ and noisy input image $x_t$, i.e.,~$p(\text{noise} \mid t, x_t)$, in order to progressively denoise the input such that a new sample $x_0$ from the target data distribution can be generated. 
(B) We have recently hypothesized that the brain performs an analogous denoising process to navigate through possible neural activity patterns in search of a refined goal state -- most likely via reinforcement learning (RL) in environments where the precise goal state is unknown to the agent. Denoising (i.e., uncertainty reduction) provides a natural candidate mechanism that the brain can apply even when the specific target state is completely unknown. In decoded neurofeedback (DecNef), the participant seeks brain states that minimize the discrepancy between the current neural pattern and the \textit{distribution} of desired target states; the degree of match is presented to the participant as a visual representation of the computed reward. 
(C) The closed-loop DecNef procedure enables human participants to learn to denoise (refine) their own brain states through reinforcement learning. Neural response patterns (BOLD signal) are acquired from a region of interest using functional magnetic resonance imaging (fMRI), compared against a predefined target neural pattern (typically derived from prior activity), and the degree of similarity between the current and target state is continuously fed back to the participant in the form of a visual feedback signal (here depicted as a circle).}
    \label{fig:introDenoisingCartoon}
\end{figure*}

An ideal paradigm to address these gaps would be one in which higher-order estimates about first-order uncertainty -- and especially about expected first-order uncertainty -- can play a measurable role in driving behavior. A promising candidate is Decoded Neurofeedback (DecNef), a real-time, noninvasive neuroimaging technique in which human participants learn via reinforcement learning to produce target neural patterns \citep{laconte2011decoding, watanabe2017advances} (see Fig.~\ref{fig:introDenoisingCartoon}). How human participants `solve' DecNef remains an open question \citep{Shibata2021}. Here, we hypothesized that participants may solve DecNef tasks by learning the distribution of task-relevant first-order representational uncertainty and minimizing it to reach a goal state \citep{gottlieb2013information,gottlieb2018towards}. %This hypothesis is motivated also by empirical evidence that individuals with higher metacognitive efficiency show superior initial performance in unconsciously selecting optimal actions based on hidden brain states during DecNef, implicating higher-order representations of uncertainty in driving DecNef performance \citep{cortese2020unconscious}.

To access and characterize these higher-order representations of expected uncertainty, here we leverage generative artificial intelligence algorithms -- specifically, denoising diffusion models designed to learn and subtract noise to reveal a target state. Building on prior work integrating AI and neuroscience \citep{cross2021using, lebel2021voxelwise, nunez2019voxelwise, nishimoto2011reconstructing, Yamins2014Performance}, we developed the Noise Estimation through Reinforcement-based Diffusion (NERD) model and trained it using previously collected functional magnetic resonance imaging (fMRI) DecNef data \citep{Cortese2021}. Importantly, NERD is trained with reinforcement learning to mimic the unsupervised manner in which the brain likely learns about its own noise \citep{azimi2024diffusion}. We then evaluate how well the model's learned noise distributions capture human behavior by comparing it to a control model trained via backpropagation. To anticipate, our findings suggest that human subjects do learn about their own expected uncertainty distributions to solve DecNef tasks, and that the NERD model presents a powerful tool for accessing and characterizing higher-order neural representations about uncertainty.

%We found that NERD explained substantively more individual variation in participants' ability to solve the task for three studies ($R_\text{NERD}^2 = 0.782, 0.76, 0.73$) than did the control model ($R_\text{control-diffusion}^2 = 0.321, 0.28, 0.08$). Crucially, this explanatory power was enhanced when we discovered clusters in the distributions of expected noise learned by NERD for each participant ($R_\text{NERD}^2 = 0.89, 0.92, 0.86$), much more so than the control model ($R_\text{control-diffusion}^2 = 0.43, 0.31, 0.12$). These findings suggest that human subjects do learn about their own expected uncertainty distributions to solve DecNef tasks, and that the NERD model presents a powerful tool for accessing and characterizing higher-order neural representations about uncertainty.

\section*{Materials and Methods}

\subsection*{Experimental Design}

{\label{212236}}
\subsubsection*{Human Neuroimaging Dataset}

We used pre-existing data from Studies 1--3 in the DecNef collection dataset \citep{Cortese2021}. In Study 1, 24 participants (4 female, 20 male; mean age 22.3 ± 2.2 years) learned to modulate cingulate cortex multivoxel patterns \citep{shibata_neurofeedback2016}. Study 2 included 12 participants (1 female, 11 male; mean age 21.2 ± 1.8 years) and examined similar modulation effects in early visual cortex \citep{amano2016learning}. Study 3 included 9 participants in the primary neurofeedback group (3 female, 6 male; mean age 23.6 ± 3.6 years) and 17 participants in a combined control-related sample (6 female, 11 male; mean age 23.5 ± 2.8 years), also targeting early visual cortex \citep{Koizumi2016}; here we used only the data from the neurofeedback group.

%Each study began with an fMRI session to collect data for constructing a multivoxel ``decoder'' (a classifier of brain-activity patterns) \citep{shibata_neurofeedback2016}. 
Each experiment comprised multiple stages: pre-test (Study 1 only), decoder construction, induction (neurofeedback), post-test, and interview. In \emph{pre-test} (Study 1 only), participants rated their preferences for face images, to identify individual-specific ``neutral'' face stimuli and because the eventual DecNef purpose would be to change these facial preferences. Pre-test was not necessary for Studies 2 and 3, because their targets were objectively defined: in Study 2 subjects had to induce multivoxel patterns of the color red in their V1/V2, and in Study 3 they had to learn to regulate visual cortex activity to achieve patterns induced by viewing a gray vertical grating.

In \emph{decoder construction}, a region-specific decoder was trained to map voxel patterns to ratings. The decoder output was:
\begin{equation}
R_{\text{decoded}} = W_{\text{voxel}}^T \cdot A_{\text{voxel}} + b
\label{eq:decnefDecoder}
\end{equation}
where \( A_{\text{voxel}} \) is the voxel activation pattern, \( W_{\text{voxel}} \) are decoder weights, and \( b \) is the mean behavioral rating. During \emph{induction} (6 s), participants were instructed to modulate brain activity to maximize subsequent feedback: decoder likelihood of the target state was displayed as a disc within a circle (disc size = reward; circle = maximum).
In Study 1, the disk size (radius) was calculated proportionally to the decoder's output: the estimated facial preference rating decoded from the cingulate cortex activation pattern during the induction period (higher rating = larger disk for the higher-preference group; opposite for the lower-preference group). Similarly, in Study 2, the disk size (radius) = the likelihood that the BOLD activation pattern in V1/V2 during the induction period matched the target color-orientation association pattern. In the same way, In Study 3, the disk size (radius) was calculated proportionally to the decoder's output: the likelihood that the BOLD activation pattern in early visual cortex during the induction period matched the target activity pattern associated with fear reduction (counter-conditioning).
Study 1 additionally presented a neutral face cue (0.5 s) before induction, whereas Studies 2--3 used an achromatic grating during induction.

%\subsubsection{Neuroimaging Data Acquisition and Preprocessing}

Whole-brain blood oxygen level dependent (BOLD) signals were acquired on 3T MRI scanners (Studies 1--2: Verio, Siemens, TR = 2 s, TE = 26 ms; Study 3: Trio, Siemens, TR = 2 s, TE = 30 ms; FA = 9°, voxel size = 3 × 3 × 3.5 mm³, matrix size = 64 × 64) \citep{shibata_neurofeedback2016}. Each participant’s dataset included ~4000 repetition times (TRs) across 12 runs (avg 216 TRs each) and 10 induction-day runs. High-resolution structural scans (T1-weighted MP-RAGE, voxel size = 1 × 1 × 1 mm³) provided anatomical reference. Data were preprocessed using BrainVoyager QX, with 3D motion correction, rigid-body co-registration, gray matter masking, 4-second hemodynamic delay adjustment, z-score normalization, and linear trend removal. No smoothing was applied. Additional information about this dataset, including neuroimaging acquisition and processing details, can be found in the original study \citep{shibata_neurofeedback2016}.

{\label{130442}}
% \section{Methods}
%Our goal is to develop the Noise Estimation via Reinforcement-based Denoising (NERD) model -- a reinforcement learning (RL)-based denoising diffusion model that learns noise-distribution higher-order representations (HORs) from functional magnetic resonance imaging (fMRI) data, mirroring human learning about its own noise \citep{asraridiffusion}. NERD transforms any random multivoxel brain pattern into a pattern close to those of a target distribution of multi-voxel patterns through iterative denoising steps, each step representing a sample from a noise-distribution HOR. While these HORs are learned in voxel space, they serve as proxies for neural representations \citep{baker2022three} which we can reveal through abstraction via dimensionality reduction, which allows us to characterize their relationship to mental structures. We train NERD models using Diffusion-Driven Policy Optimization (DDPO) \citep{wang2023ddpo, black2023training} and compare their performance to that of control-diffusion models trained via backpropagation. Below, we detail the fMRI dataset, model architectures, training procedures, and evaluation metrics.

\subsection*{Denoising Diffusion Models}

\subsubsection*{Overview and Introduction to Denoising Diffusion Models}

Denoising diffusion models, also referred to as diffusion probabilistic models, have emerged as a leading class of generative models renowned for their stable training dynamics and high-fidelity sample quality \citep{ho2020denoising,dhariwal2021diffusion}. Inspired by non-equilibrium thermodynamics, these models generate data by learning to reverse a gradual `noising' process that progressively corrupts clean samples into pure Gaussian noise over multiple timesteps \citep{ho2020denoising}. A neural network is trained to predict and remove the added noise at each step, enabling the generation of new samples by starting from random noise and iteratively denoising them. Compared to earlier generative approaches such as generative adversarial networks and variational autoencoders, diffusion models offer superior training stability ~\citep{dhariwal2021diffusion,nichol2021improved}. Denoising diffusion models are established as the state-of-the-art in text-to-image synthesis, audio generation, video modeling, and various other domains such as molecular design and weather forecasting \citep{song2020denoising,rombach2022high}.

%\subsubsection*{Introduction to Denoising Diffusion Models}

% Denoising diffusion models generate data samples by reversing Gaussian noise added to original samples \citep{ho2020denoising}, thereby excelling in conditional generation for applications like image synthesis \citep{gao2024bayesian, chen2024overview, peng2023medical} and outperforming generative adversarial networks \citep{dhariwal2021diffusion}. 
Training diffusion models involves first adding (typically Gaussian) noise over \( T \) steps to an input \( x_0 \), forming a sequence \( x_1, \dots, x_T \):
$q(x_t | x_{t-1}) = N(x_t; \sqrt{1 - \beta_t} x_{t-1}, \beta_t I)$
where \( \beta_t \) controls noise magnitude \citep{sohl2015deep}. In traditional applications, the denoising model then learns the reverse process \( p_\theta(x_{t-1} | x_t) \), minimizing Kullback-Leibler divergence between the learned reverse transitions and the true denoising steps via the evidence lower bound on the data likelihood \citep{kingma2021variational}. Importantly for our goals, recent approaches have simplified this training objective by reparameterizing the diffusion process, allowing the model to learn and thus predict the added noise $\epsilon$ directly. This reparameterization results in the simplified objective function: $L_{\text{simple}} = \mathbb{E}_{x_0, \epsilon, t} \left[ | \epsilon - \epsilon_\theta(x_t, t) |^2 \right]$ where \( \epsilon_\theta \) is the model’s noise prediction \citep{ho2020denoising, song2021scorebased}.

\subsubsection*{Noise Estimation through Reinforcement-based Diffusion (NERD) Model}

The NERD model is an adaptation of diffusion models for modeling the noise-learning process in DecNef using a reinforcement learning framework based on diffusion-driven policy optimization \citep{black2023training}. NERD maximizes the same reward function used to train human participants, \( R(x) \) (Eq.~\ref{eq:decnefDecoder}), aligning generated brain states with neurofeedback targets. A policy network \( \pi_\theta(x_t | s_{t-1}) \), with state \( s_t = (c, t, x_t) \) (context \( c \): maximize feedback; timestep \( t \); noisy sample \( x_t \)), optimizes $\mathbb{E}_{x \sim \pi_\theta} [R(x)]$.
The hybrid loss function thus combines the diffusion and reinforcement learning objectives:
\vspace{-1mm}
\begin{equation}
L_{\text{NERD}}(\theta) = \mathbb{E}_{t, x_0, \epsilon} \left[ \| \epsilon - \epsilon_\theta(x_t, t) \|^2 \right] - \lambda \mathbb{E}_{x \sim \pi_\theta} [R(x)]
\label{eq:RLloss}
\end{equation}
where \( \lambda \) is a balancing factor that scales the reinforcement learning reward component relative to the diffusion model's denoising loss \citep{black2023training}. We modeled denoising as a Markov Decision Process, with state \( s_t \), action \( a_t \) (generating \( x_{t-1} \)), and reward \( R(x) \) at the final step \( x_0 \). Finally, we used the REINFORCE algorithm \citep{williams1992simple} to optimize \( \theta \), computing cumulative reward $G_t = \sum_{k=0}^{T-t} \gamma^k r_{t+k+1}$ where \(\gamma\) \((0 < \gamma < 1)\) is the discount factor (emphasizing the importance of near-term rewards while still accounting for long-term outcomes). The return \(G_t\) represents the total reward the agent expects to accumulate from time \(t\) onward and is used to guide parameter updates.

The REINFORCE algorithm updates the policy parameters \(\theta\) using the likelihood ratio policy gradient. The gradient of the expected reward with respect to \(\theta\) is given by $\nabla_{\theta} J(\theta) = \mathbb{E}_{\pi} \left[ \sum_{t=1}^T G_t \nabla_{\theta} \log \pi_{\theta}(a_t | s_t) \right]$, where \(J(\theta)\) is the objective function representing expected reward. This gradient provides the direction in parameter space that maximizes expected cumulative reward. In practice, this expectation is approximated through sampled episodes, resulting in the following update rule: \vspace{-1mm}
\begin{equation}
   \theta \leftarrow \theta + \alpha \sum_{t=1}^T G_t \nabla_{\theta} \log \pi_{\theta}(a_t | s_t)
\end{equation}
where \(\alpha\) is the learning rate that controls the update step size. We stabilized training through gradient clipping (to prevent exploding gradients) and through employing a constant baseline function \(b(s_t) = \mathbb{E}[G_t]\), subtracted from \(G_t\) in the update rule, to reduce the policy gradient estimate variance without introducing bias \citep{sutton2018reinforcement}. We updated the parameters of the policy model in batches of 32 episodes, defined as `training epochs'.

\subsubsection*{Control Model}

The control-diffusion model, identical in architecture to NERD, was trained via backpropagation \citep{rumelhart1986learning} to maximize reward deterministically. Removing sampling from the learning process simplifies the training dynamics while preserving the generative properties of the diffusion model and reward-driven supervision. The control-diffusion model's loss function is:
\vspace{-1mm}
\begin{equation}
L_{\text{control}}(\theta) = -R(x)
\label{eq:controlloss}
\end{equation}
Gradients were again computed over training epochs. 

\subsubsection*{Policy Network Architecture}

\begin{figure}[ht]
    \centering
    {\includegraphics[width=.8\textwidth]{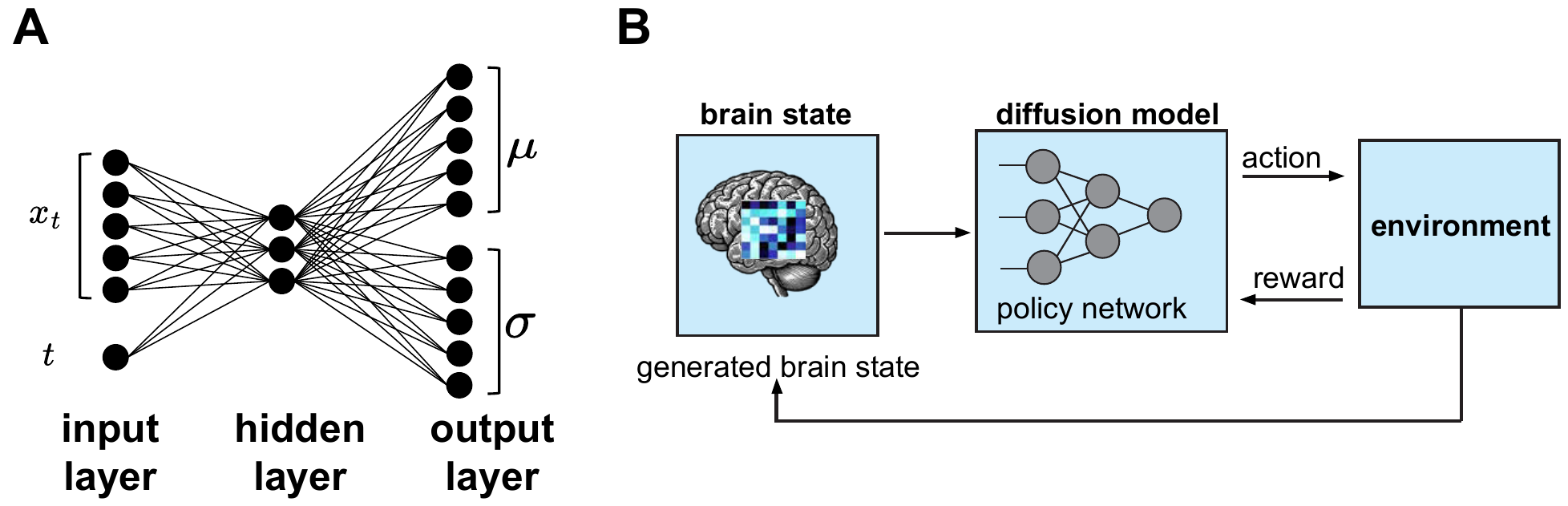}}
    \caption{\textbf{Policy network architecture and closed-loop training.} (A) The network has an input layer (state dimension + 1), a 128-node hidden layer, and an output layer (2 × state size) estimating \( \mu \) and \( \sigma \) for the learned noise distribution at each denoising step, i.e. \( p(\mu, \sigma | t, x_t) \). (B) Brain state (TR) is input, denoised, and passed to the decoder (Eq.~\ref{eq:decnefDecoder}) for feedback, leading to updates to the parameters \( \theta \) of the policy network.}
    \label{fig:policy_network}
\end{figure}

Both model families (NERD and control-diffusion) used a fully connected neural network \citep{williams1992simple} with an input layer (state dimension + 1), a 128-node hidden layer, and an output layer (2 × state size) estimating mean \( \mu \) and standard deviation \( \sigma \) of the learned noise distribution at each denoising timestep, \( p(\mu, \sigma | t, x_t) \sim N(\mu, \sigma)\) (Fig.~\ref{fig:policy_network}), which is assumed to be Gaussian following standard convention in diffusion modeling \citep{ho2020denoising} and the proliferation of Gaussian noise assumptions in neuroscience applications \citep{dewit2019information,jonsdottir2013levy, mausfeld2012cognitive, parker2022assumptions}. This setup allows the network to output the estimated parameters for the policy distribution over actions, enabling the model to estimate the noise (i.e., the position within the noise distribution) at each denoising time step \({t}\).

\subsection*{Model Training}

For each subject in each DecNef study, we trained a paired set of subject-specific models (one NERD, one control-diffusion) using all collected volumes (repetition times; TRs). In a closed-loop framework (Fig.~\ref{fig:policy_network}b), each model received a region of interest voxel-space state, performed denoising, and received feedback via the pre-trained decoder (Eq.~\ref{eq:decnefDecoder}). Feedback was used to update the policy network parameters, and the updated state was re-input to the model for the next iteration.

\subsection*{Evaluation Metrics}

\subsubsection*{Model Learning Assessment}

We assessed the NERD and control-diffusion models’ ability to maximize reward (Eq.~\ref{eq:decnefDecoder}) and minimize loss (Eqs.~\ref{eq:RLloss}, \ref{eq:controlloss}) across 40 denoising steps, evaluating final states after all denoising steps \( x_0 \).

\subsubsection*{Fitting Models to Human Data}

To avoid over-training, we identified the training epoch which minimized the Negative Log Likelihood (NLL) of the model parameters given the human data for each participant:
\vspace{-1mm}
\begin{equation}
NLL_{m,e,trial,subject} = -\log p(x_{subject,trial} | X_{0,m,e})
\label{eq:NLL}
\end{equation}
where \( X_{0,m,e} \) is the distribution of 30 generated states per trial (each produced in response to a single input state corresponding to \( x_{subject,trial} \)), and \( x_{subject,trial} \) is the human voxel pattern for that trial. We defined the best-fitting epoch for each subject as:
\vspace{-1mm}
\begin{equation}
e^*_{m,subject} = \min_e \left( \frac{1}{n} \sum_{trial}^{n} NLL_{m,e,trial,subject} \right)
\label{eq:minNLL}
\end{equation}

\noindent where m is the the model's index. Models were frozen at \( e^* \) for each subject for further analyses. 

\subsubsection*{Extracting Higher-order Expectations about Noise}

We first computed mean reward \( R_{m,t} \) achieved per denoising step across all subjects for both model families to evaluate their ability to learn the denoising task.

% Next, we examined the internal representations of noise distributions that both model families learned by calculating the similarity (Pearson correlation) between (a) pairs of patterns produced at each denoising step for each input TR, and (b) pairs of trials from the DecNef experiment:
% \vspace{-1mm}
% \begin{equation}
%     r = \frac{\sum_{i=1}^{n} (x_i - \bar{x})(y_i - \bar{y})}{\sqrt{\sum_{i=1}^{n} (x_i - \bar{x})^2  (y_i - \bar{y})^2}}
%     \label{eq:pearsoncoefficient}
% \end{equation}

Next, recall that our ultimate goal is to examine the higher-order expectations about noise themselves, i.e. ~$p(\mu, \sigma | t, x_t)$.
%p(noise|step,input_{step})$. 
We therefore extracted the learned parameters \(\mu\) and \(\sigma\), normalized these between 0 and 1 for each voxel to aid visualization ($\mu_{vt}^* = \frac{\mu_{vt}-\min\limits_{t}(\mu_{vt})}{\max\limits_{t}(\mu_{vt})-\min\limits_{t}(\mu_{vt})}$ and $\sigma_{vt}^* = \frac{\sigma_{vt} - \min\limits_{t}(\sigma_{vt})}{\max\limits_{t}(\sigma_{vt}) - \min\limits_{t}(\sigma_{vt})}$), and then clustered them using K-means implemented in \texttt{scikit-learn} \citep{Pedregosa2011ScikitLearn} to visually inspect patterns. We then employed multi-step Principal Components Analysis (PCA) \citep{Pearson1901, Bishop2006} to reduce the [\( \mu \), \( \sigma \)] trajectories across all denoising steps to three dimensions, to abstract from voxel space to representation or mental space. Recall that, in the studies we used from the DecNef Collection, the mental structure in question consists of first-order representations -- e.g., about faces in Study 1, presumably with many dimensions beyond preference/attractiveness (e.g., distinctiveness, memorability, familiarity, identity, expression, and many more \citep{hancock1996face,rhodes2015distinct}). Thus, in Study 1, the task-relevant first-order representation dimension is `preference'; likewise, for Studies 2 and 3 it is color space or other similar features related to encoding a gray vertically-oriented grating. Thus, this PCA step allows us to discover the \textit{direct mapping} between the model's learned noise representations in voxel space (i.e., $p(\mu,\sigma | t,x_t)$) and those in first order representation space along these task-relevant dimensions. PCA was done in two stages: first to $\mu$ and $\sigma$ together across denoising steps $t$ for each voxel to characterize a single dimension capturing variance fo that voxel, and then using this single dimension to all voxels collectively to characterize how they collectively move through this reduced feature space. 

To align the trajectories across subjects and enable direct comparison in the reduced-dimensional space, we applied orthogonal Procrustes analysis \citep{Schonemann1966}. The trajectory of the first subject was chosen as the reference, $Y_\text{ref} \in \mathbb{R}^{T \times 3}$ (across $T$ denoising steps in the 3D principal component space). Each subsequent subject's trajectory $Y$ (in the same space) was then rigidly aligned to this reference by computing the optimal rotation matrix $R$ that minimizes the Frobenius norm of the difference, i.e. minimizes $\| Y R - Y_{\text{ref}} \|_F$. The aligned trajectory was then calculated as $Y' = Y R$.
Following alignment, we computed the mean residual displacement of each aligned trajectory relative to the reference as
\vspace{-1mm}
\begin{equation}
d = \frac{1}{T} \sum_{t=1}^{T} \bigl\| y'_t - \bar{y}_{\text{ref}} \bigr\|_2
\end{equation}
where $y'_t$ is the aligned point at denoising step $t$ and $\bar{y}_{\text{ref}}$ is the centroid (time-averaged position) of the reference trajectory. This Procrustes alignment step removed rotational degrees of freedom, ensuring that subsequent between-subject dissimilarities reflected meaningful differences in denoising dynamics rather than arbitrary orientation in the reduced feature space. Finally, we computed representational dissimilarity matrices containing the pairwise Euclidean norm distances ($||x|| := \sqrt{x\cdot x}$) between subjects' aligned trajectories through this reduced-dimensionality noise-distribution higher-order space, and then used hierarchical clustering to identify groups of participants who exhibited similar denoising trajectories (indicating they had learned and were navigating similar higher-order expectations about noise).

\subsection*{Statistical Analysis}

{\label{417318}}

After freezing the models at \( e^* \), we first evaluated the degree to which each model family could predict individual differences in the reward (Eq. \ref{eq:decnefDecoder}) achieved by the human subjects. For each model family, we fitted a simple linear model of the form \texttt{y $\sim$ x} to predict the actual mean reward achieved by each subject across all DecNef trials from the mean reward predicted for that human subject by the subject's best-fitting model. 

We subsequently assessed whether cluster membership from the dimensionality reduction, trajectory alignment, and clustering analysis -- indicating similarity between participants in their learned distributions of expected noise -- could further explain individual differences in DecNef performance. This involved including a second categorical predictor variable in the linear models to capture potential differences in this relationship as a function of cluster, \texttt{y $\sim$ x1 * x2}, where \texttt{x1} is the continuous variable of mean model predicted reward, as before, and we add \texttt{x2} as the categorical variable of cluster. We included the interaction term to allow for differences in possible predictive power as a function of cluster.

For both sets of linear models (with and without the cluster factor), we evaluated fitted parameters and goodness of fit metrics for each model family (NERD and control-diffusion).

\subsection*{Implementation}

All models and analyses were implemented with custom scripts in Python 3.12.3 using PyTorch 2.3.1.

\section*{Results}

{\label{502277}}

%Our goal was to develop the Noise Estimation via Reinforcement-based Denoising (NERD) model -- a reinforcement learning-based denoising diffusion model that learns higher-order representations of expected uncertainty from functional magnetic resonance imaging (fMRI) data, mirroring human learning about its own noise \citep{asraridiffusion}. NERD transforms any input multivoxel brain pattern into a pattern close to those representing a target distribution of multi-voxel patterns through iterative denoising steps, each step representing a sample from the learned distribution of expected noise in the fMRI data. While these expectations about noise are learned in voxel space, they serve as proxies for neural representations \citep{baker2022three} which we can reveal through abstraction via dimensionality reduction, which allows us to characterize their relationship to mental structures. We trained NERD models -- one for each participant -- using Diffusion-Driven Policy Optimization (DDPO) \citep{wang2023ddpo, black2023training} on three pre-existing DecNef fMRI datasets \citep{Cortese2021,shibata_neurofeedback2016,amano2016learning, Koizumi2016} and compared their performance to that of control-diffusion models trained via backpropagation on the same data. In total, we trained one pair of  subject-specific models (one NERD, one control-diffusion) using the existing fMRI DecNef data collection \citep{Cortese2021}. See Methods for details of model training, fitting, and evaluation metrics. 

%\subsection{Fitting Models to Human Data}

\begin{figure}[h]
    \centering
    \includegraphics[width=0.65\textwidth]{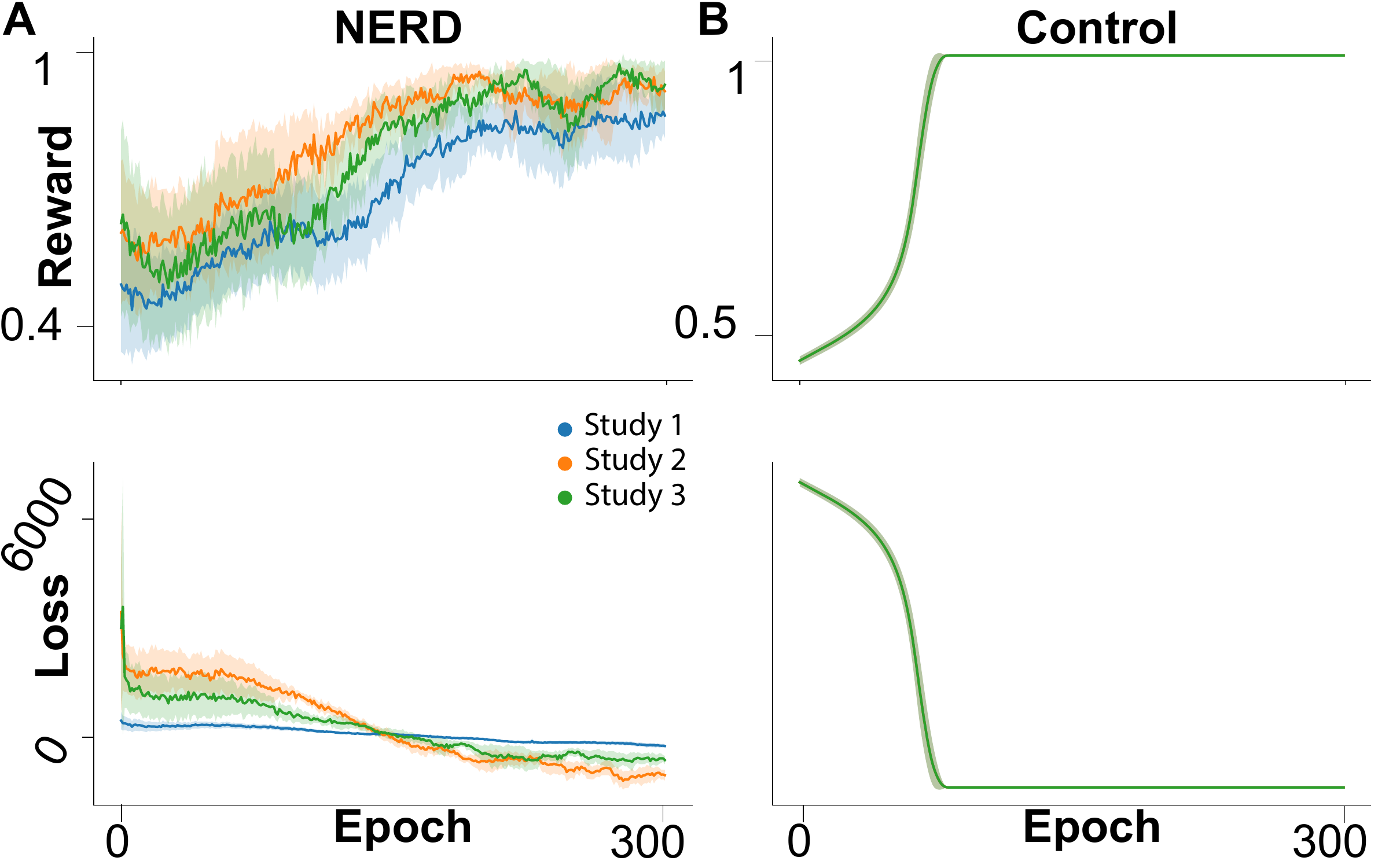}
    \caption{\textbf{Model learning progress.} 
    (A) NERD and (B) control-diffusion models show decreased loss (top) 
    and increased reward (bottom) across training epochs.}
    \label{fig:performance}
\end{figure}

We first wanted to establish that both groups of models could learn the denoising task by examining general reward and loss trajectories as a function of training epoch. Both the NERD and control-diffusion models reduced loss and increased cumulative reward (Eq.~\ref{eq:decnefDecoder}) for all subjects as training progressed across epochs, indicating effective learning of the DecNef task policy (Fig. ~\ref{fig:performance}). 

\begin{figure}[t]
    \centering
    \includegraphics[width=.8\textwidth]{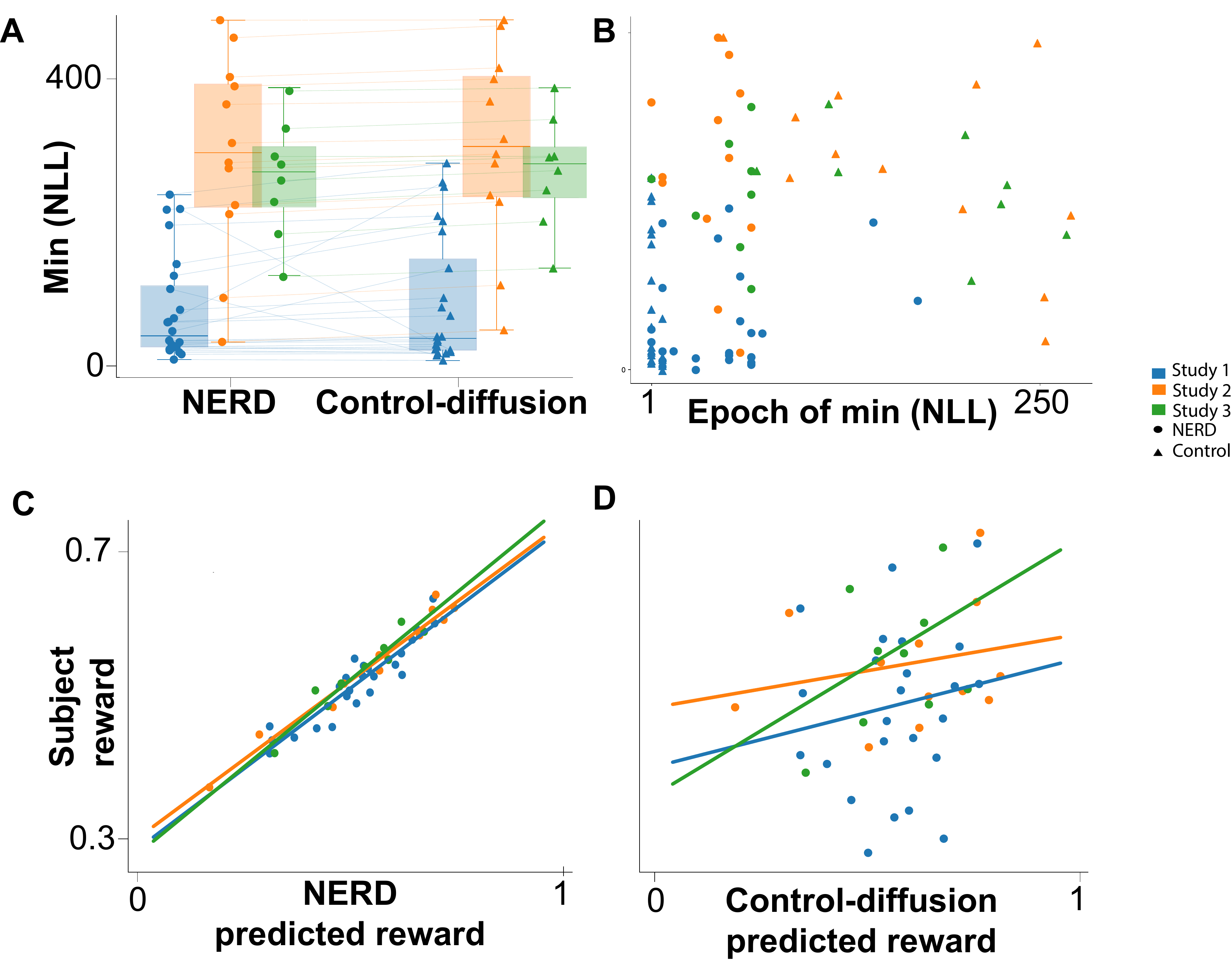}
    \caption{\textbf{Model fitting results.} (A) Both NERD and control-diffusion models predicted humans' multivoxel patterns equally well: $min(NLL)$ distributions show no difference between models. (B) NERD models reached min(NLL) at later epochs (mean $e^*_\text{NERD}$ for Studies 1--3: = 145.6 $\pm$ 32.2, 128.6 $\pm$ 28.3, 36.4 $\pm$ 11.2; $e^*_\text{control-diffusion}$ for Studies 1--3: 4.8 $\pm$ 1.8, 39.2 $\pm$ 8.5, 48.7 $\pm$ 9.1). (C,D) Linear models revealed that NERD models ($R^2$ for Studies 1--3: 0.782, 0.76, 0.73) out-performed than control-diffusion models ($R^2$ for Studies 1--3: 0.321, 0.28, 0.08) in predicting human rewards from model rewards despite no difference in goodness of fit ($min(NLL)$).}
    \label{fig:nll}
\end{figure}

Because both the NERD and control models could ultimately learn the task much better than the human subjects in the DecNef experiment, we next sought to freeze the models at a training epoch that maximized their ability to produce distributions of voxel patterns that best matched the same capacity actually achieved by the human subjects. Thus, we determined the best-fitting model for each participant by minimizing Negative Log Likelihood (NLL, Eq.~\ref{eq:NLL}) of model-predicted patterns $\text{w.r.t.}$ human voxel patterns during DecNef induction trials (see Methods). Both model families achieved similar NLL for all three studies (Fig.~\ref{fig:nll}a), but NERD models required more epochs to reach minimal NLL (mean $e^*_\text{NERD}$ = for studies 1--3: 134.5 $\pm$ 32.4, 128.6 $\pm$ 28.3, 39.2 $\pm$ 8.5; $e^*_\text{control-diffusion}$ for studies 1--3: 36.4 $\pm$ 11.2, 39.2 $\pm$ 8.5, 48.7 $\pm$ 9.1; Fig.~\ref{fig:nll}b). The epoch of minimal NLL ($e^*$) was uncorrelated with NLL (Table~\ref{tab:correlations}), indicating that we successfully halted training at the optimal epoch for maximizing goodness of fit: models that were allowed to train for longer did not automatically exhibit better goodness of fit with human subjects. These results show that both NERD and the control model were equivalently able to learn distributions of target voxel patterns, such that any differences in their ability to reproduce subject rewards are not due to general differences in model training success.

\begin{table*}[t]
\centering
\begin{tabular}{l l c c}
\hline
Model & Study & $R$ & $p$ \\
\hline
\multirow{3}{*}{NERD} & Study 1 & 0.09 & 0.66 \\
& Study 2 & 0.07 & 0.72 \\
& Study 3 & -0.06 & 0.69 \\
\hline
\multirow{3}{*}{Control-diffusion} & Study 1 & -0.02 & 0.91 \\
& Study 2 & 0.04 & 0.88 \\
& Study 3 & -0.01 & 0.93 \\
\hline
\end{tabular}
\caption{Correlation coefficients ($R$) and p-values ($p$) showing lack of correlation between $e^*$ and NLL across different models and studies.}
\label{tab:correlations}
\end{table*}

Our next question was whether it matters that NERD learned these voxel pattern distributions through reinforcement learning -- i.e., how brains would likely learn these distributions -- rather than traditional backpropagation. To answer this question, we used linear models (\texttt{y $\sim$ x}) to assess how well the best-fitting models (frozen at $e^*$ per subject) were able to predict human mean DecNef rewards (across all DecNef trials) from model-predicted rewards (Fig.~\ref{fig:nll}c,d). Importantly, despite similarities in NLL between NERD and control-diffusion models -- indicating no substantive differences in the models' ability to learn target multivoxel patterns -- NERD models explained more individual variance in predicted reward (for all three studies: $  R_\text{NERD}^2 = 0.782, 0.76, 0.73 $; see Table \ref{tab:models} for $\beta$ values; all $  p < 0.001  $) than did control-diffusion models ($  R_\text{control-diffusion}^2 = 0.321, 0.28, 0.08 $; see Table \ref{tab:models} for $\beta$ and $  p  $ values). These findings suggest NERD has captured something meaningful about the way human subjects learned the DecNef task, such that use of reinforcement learning to train diffusion models may be a key component in extracting and characterizing higher-order representations about uncertainty.

\begin{table*}[t]
\centering
\begin{tabular}{l l c c c c c}
\hline
Model & Study & $R^2$ & $\beta_1$ & $p(\beta_1)$ & $\beta_0$ & $p(\beta_0)$ \\
\hline
\multirow{3}{*}{NERD} & Study 1 & 0.782 & 0.378 & $<0.001$ & 0.512 & $<0.001$ \\
& Study 2 & 0.763 & 0.362 & $<0.001$ & 0.498 & $<0.001$ \\
& Study 3 & 0.73 & 0.341 & $<0.001$ & 0.476 & $<0.001$ \\
\hline
\multirow{3}{*}{Control-diffusion} & Study 1 & 0.321 & 0.189 & 0.004 & 0.302 & $<0.001$ \\
& Study 2 & 0.284 & 0.172 & 0.012 & 0.287 & $<0.001$ \\
& Study 3 & 0.08 & 0.095 & 0.118 & 0.1456 & 0.041 \\
\hline
\end{tabular}
\caption{Regression results comparing the variance explained in predicted reward values. 
NERD models consistently outperformed control-diffusion models across all three studies, 
showing higher $R^2$ values and significant regression coefficients 
($\beta_0$ and $\beta_1$, all $p < 0.001$). Control-diffusion models exhibited 
substantially lower explanatory power, especially in Study 3.}
\label{tab:models}
\end{table*}

%\subsection{Reward Trajectories Through Denoising}

\subsection*{Higher-order Expectations about Noise}

\begin{figure}[htbp]
    \centering
    {\includegraphics[width=.6\textwidth]{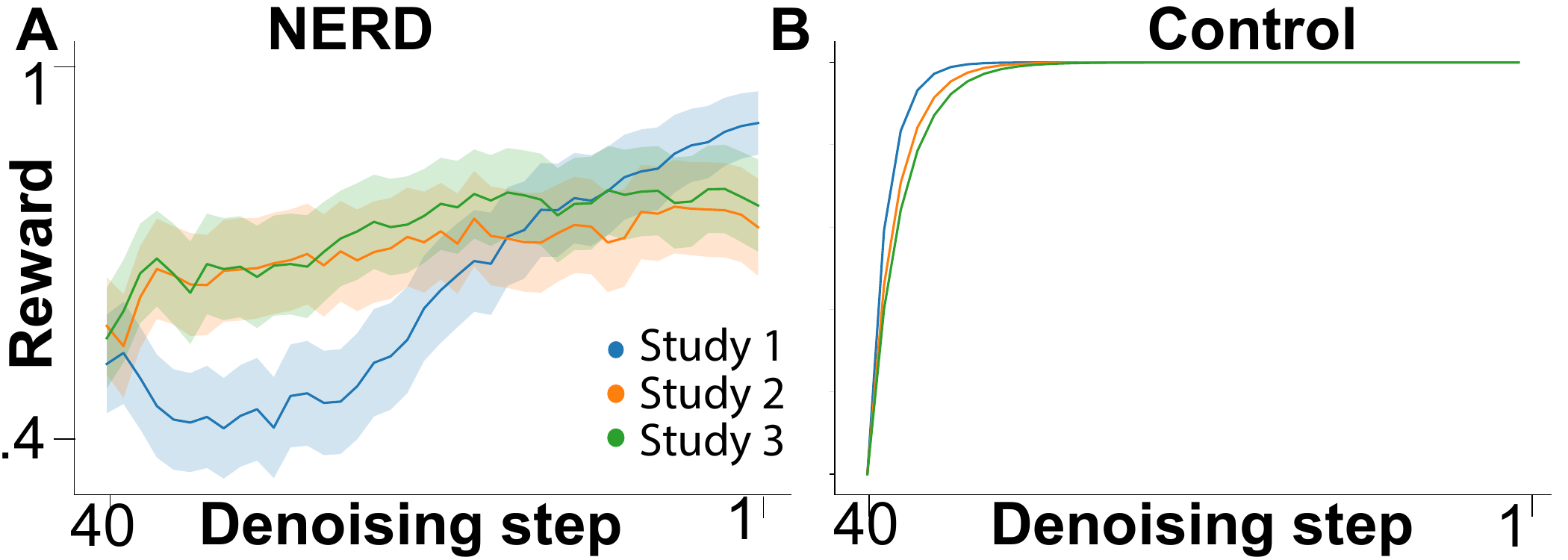}}
    \caption{\textbf{Reward trajectories.} (A) NERD models show gradual reward increase as denoising progresses. (B) Control-diffusion models achieve maximal reward nearly immediately. Solid lines show mean reward; shaded areas show standard deviation.}
    \label{fig:rewardThruDenoising}
\end{figure}

Finally, we can begin to analyze the learned noise distributions themselves. One clue about differences in these distributions between NERD and control models would be in their dynamics, which we can superficially access through examining the accumulation of reward across the 40 denoising steps. Visualization of these trajectories revealed that NERD accumulated reward gradually (Fig.~\ref{fig:rewardThruDenoising}a), while control-diffusion models maximized reward abruptly (Fig.~\ref{fig:rewardThruDenoising}b), reflecting distinct denoising dynamics learned by the two model families.

\begin{figure*}[ht]
    \centering
    {\includegraphics[width=1\textwidth]{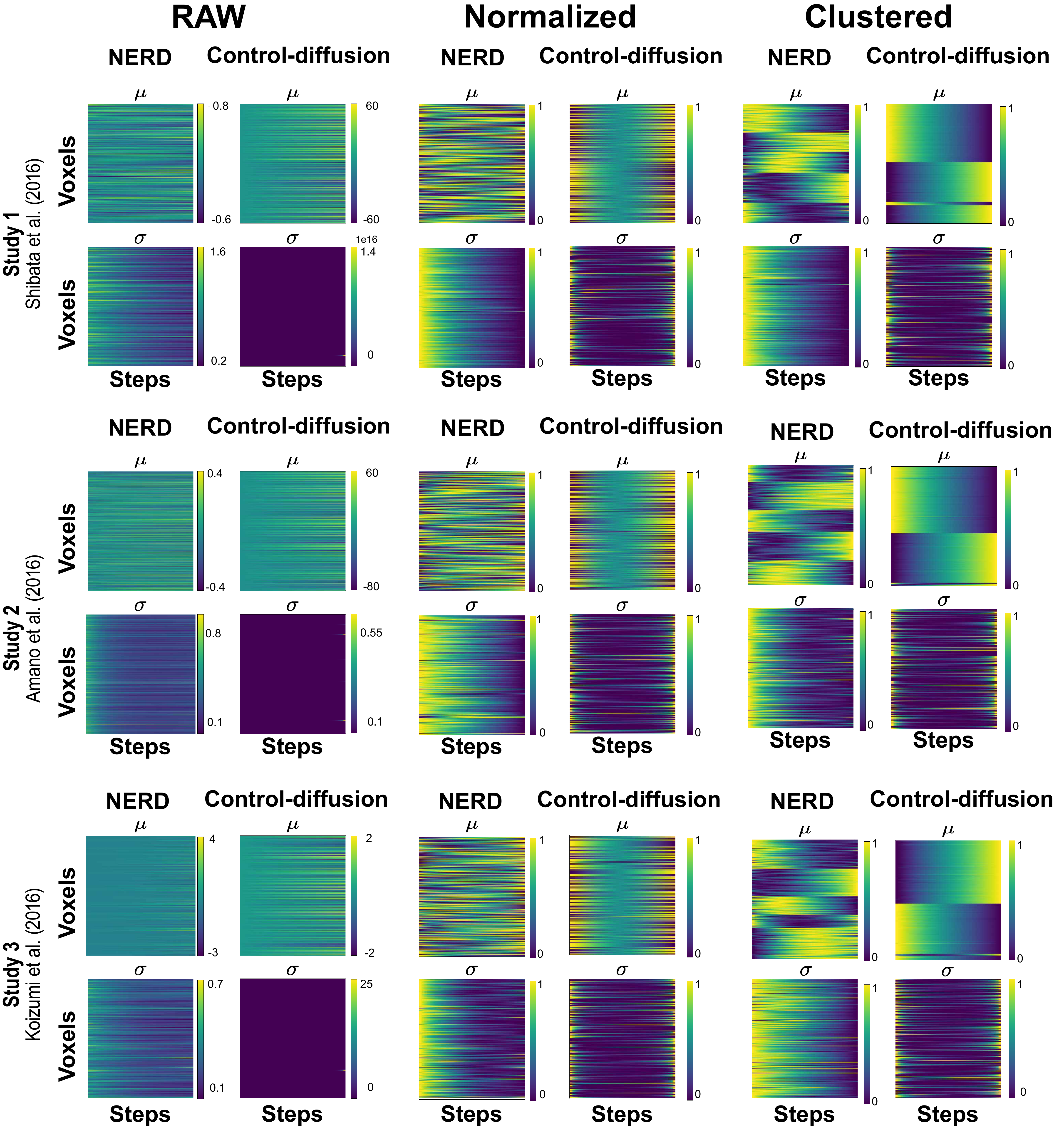}}
\caption{\textbf{Raw, normalized, and clustered trajectories of $\mu$ and $\sigma$ estimated by the models.} [\( \mu \), \( \sigma \)] estimates (raw: left column; normalized: middle column; clustered: right column) for three representative participants -- one for each study. NERD models show heterogeneous, non-monotonic \( \mu \) and decreasing \( \sigma \); control-diffusion models show monotonic \( \mu \) and mixed \( \sigma \). Each row shows average results for each one of the three studies}
    \label{fig:noiseTrajectories}
\end{figure*}

Ultimately, however, our goal is to learn about the nature of higher-order expected noise distributions, not just how they behave. We thus next turned to directly examining learned noise distributions \( p(\mu, \sigma | t, x_t) \) in voxel space, which map to the task-relevant first-order representational dimension (facial preference, color-orientation association, or induction of fear-associated neural patterns, depending on study). As a reminder, these are the mean \( \mu \) and variance \( \sigma \) of the noise learned across all denoising steps for each voxel (see Methods). For basic visualization we plotted these raw values for three representative participants (Figure~\ref{fig:noiseTrajectories}a, left column), then normalized them such that each voxel's minimum and maximum \( \mu \) and \( \sigma \) were 0 and 1 (Figure~\ref{fig:noiseTrajectories}a, middle column); finally, we clustered voxels by their similarity in the trajectory of \( \mu \) and \( \sigma \) across denoising steps (Figure~\ref{fig:noiseTrajectories}a, right column). NERD models displayed heterogeneous \( \mu \) patterns, including non-monotonic clusters where \( \mu \) first increased and then decreased or vice versa, and mostly decreasing \( \sigma \) which may indicate iterative refinement of estimated noise distributions. In contrast, control-diffusion models showed only monotonic \( \mu \) changes and mixed \( \sigma \) trends with steeper gradients. Though we plot three representative subjects, we performed these steps for all subjects in all three studies, with results highly similar to those visualized in Figure \ref{fig:noiseTrajectories}.

\begin{figure}[t]
    \centering
    {\includegraphics[width=.55\textwidth]{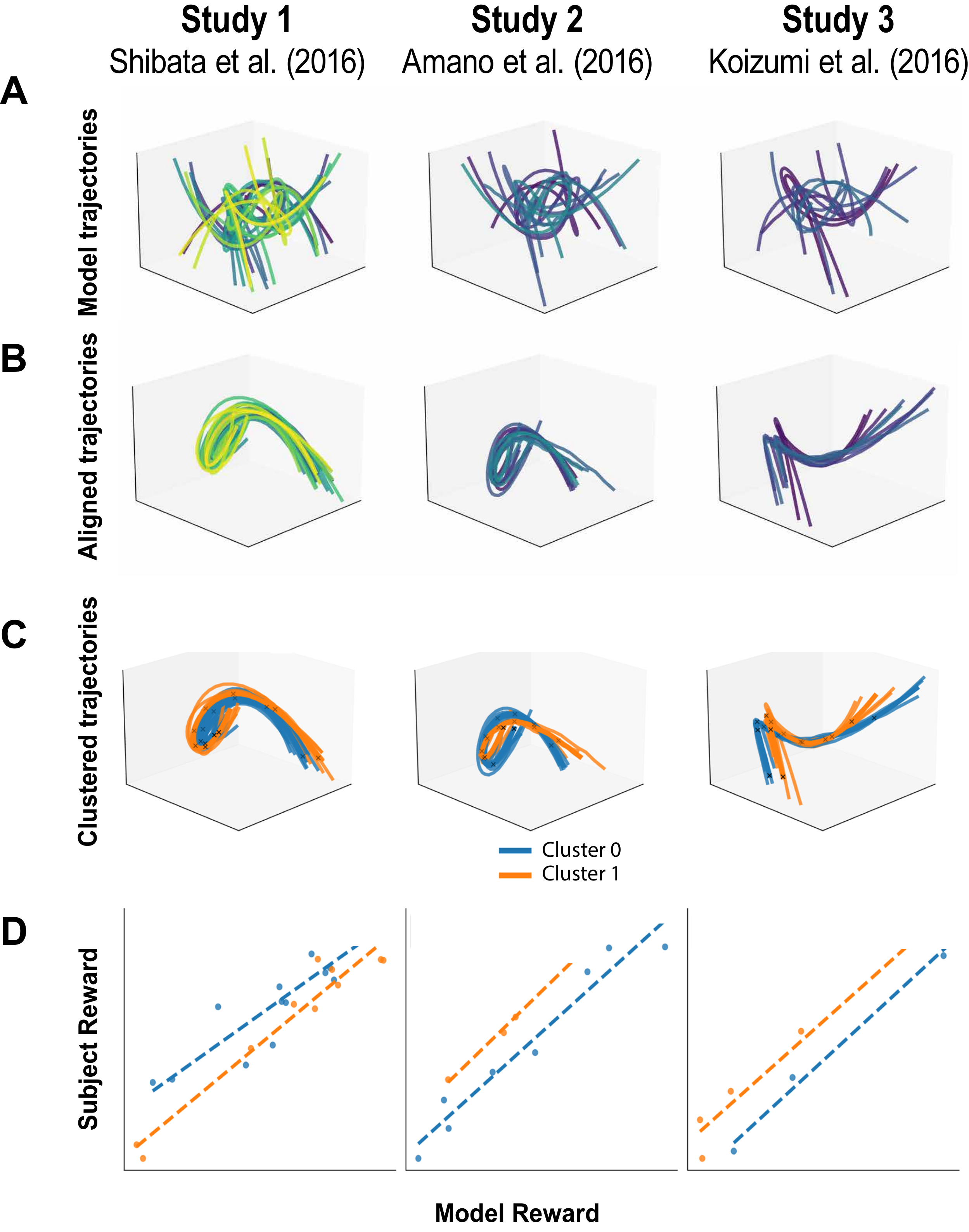}}
    \caption{\textbf{Noise distributions and state-space analysis.} (A) Raw trajectories for all three studies. (B) Trajectories after orthogonal Procrustes alignment. (C) Clustered aligned trajectories (Cluster 0: blue, Cluster 1: orange). (D) Linear models show the clusters outperform original linear models  ($R^2 = 0.89, 0.92, 0.86$) in predicting human behavior.}
    \label{fig:noiseDistributions}
\end{figure}

We then used two-stage PCA on the [\( \mu \), \( \sigma \)] trajectories to transform the learned noise distributions from voxel space into mental representational space \citep{Bishop2006, cunningham2014dimensionality, jazayeri2021interpreting,pang2016dimensionality,Pearson1901} (Fig.~\ref{fig:noiseDistributions}a). This PCA step revealed higher effective dimensionality in NERD than control-diffusion (Table~\ref{tab:pca-results}). Crucially, aligned trajectories in this PC space (Fig.~\ref{fig:noiseDistributions}b) showed participant clusters, identified via hierarchical clustering on RDMs of denoising trajectory similarity in PC space (Euclidean distance; Fig.~\ref{fig:noiseDistributions}c). These clusters also differed between the NERD and control-diffusion model families, such that two participants clustered together under NERD would not necessarily be clustered together under the control model. 

\begin{table*}[t]
\centering
\begin{tabular}{l l c c c}
\hline
Model & Study & PC1 (\%) & PC2 (\%) & PC3 (\%) \\
\hline
\multirow{3}{*}{NERD} & Study 1 & $85.7 \pm 4.5$ & $12.2 \pm 3.8$ & $2.1 \pm 0.9$ \\
& Study 2 & $87.2 \pm 4.1$ & $10.8 \pm 3.5$ & $2.0 \pm 0.8$ \\
& Study 3 & $83.9 \pm 5.2$ & $13.5 \pm 4.3$ & $2.6 \pm 1.1$ \\
\hline
\multirow{3}{*}{Control-diffusion} & Study 1 & $94.0 \pm 1.0$ & $5.3 \pm 0.9$ & $0.7 \pm 0.2$ \\
& Study 2 & $93.5 \pm 1.2$ & $5.6 \pm 1.0$ & $0.9 \pm 0.3$ \\
& Study 3 & $94.8 \pm 0.8$ & $4.7 \pm 0.7$ & $0.5 \pm 0.1$ \\
\hline
\end{tabular}
\caption{Principal component analysis (PCA) results showing the percentage of variance explained by the first three principal components (PC1--PC3) in NERD and control-diffusion models across studies. Values are reported as mean $\pm$ standard deviation.}
\label{tab:pca-results}
\end{table*}

The critical question then is whether the expected noise distributions learned by NERD explain how well participants are able to solve the DecNef task. If people solve DecNef by learning about task-relevant variance in first-order representations, and if NERD is indeed capable of revealing meaningful individual differences in these learned noise distributions, NERD's clusters in dimensionality-reduced [\( \mu \), \( \sigma \)] space ought to explain meaningful variation in participants' DecNef abilities, and to do so better than the control model.

We thus again used linear models (\texttt{y $\sim$ x1 * x2}) to predict human rewards (\texttt{y}) from model rewards (\texttt{x1}), but now including the factor of cluster for each model family (\texttt{x2}; Fig.~\ref{fig:noiseDistributions}d). 
As we hypothesized, we found that NERD's explanatory power was substantively enhanced through the addition of this clustering variable (for all three studies: $R_\text{NERD}^2 = 0.89, 0.92, 0.86$, versus $R_\text{NERD}^2 = 0.782, 0.76, 0.73$ without clusters). While the control-diffusion model also improved with the addition of clusters (for all three studies: $R_\text{control-diffusion}^2 = 0.43, 0.31, 0.12$ versus $R_\text{control-diffusion}^2 = 0.321, 0.28, 0.08$ without clusters; see also Table~\ref{tab:linear_model_results}), NERD models' explanatory power still far outstripped control models' performance. These findings suggest that human subjects do learn about their own expected uncertainty distributions to solve DecNef tasks, and that the NERD model presents a powerful tool for accessing and characterizing higher-order neural representations about uncertainty.

\begin{table}[h]
\centering
\begin{tabular}{lccc}
\hline
 & \textbf{$n$} & \textbf{$m$ (Slope)} & \textbf{$R^2$} \\ \hline
\multicolumn{4}{l}{\textbf{Study 1}} \\
Cluster 0 & 14 & 0.830 & 0.968 \\
Cluster 1 & 10 & 0.946 & 0.946 \\
\multicolumn{4}{l}{\textbf{Study 2}} \\
Cluster 0 & 8 & 0.836 & 0.946 \\
Cluster 1 & 4 & 0.899 & 0.996 \\
\multicolumn{4}{l}{\textbf{Study 3}} \\
Cluster 0 & 4 & 0.969 & 0.992 \\
Cluster 1 & 5 & 0.940 & 0.961 \\ \hline
\end{tabular}
\caption{Linear model results predicting human DecNef rewards from model rewards and clusters. ($n$=sample size, $m$=slope).}
\label{tab:linear_model_results}
\end{table}
%As we hypothesized, NERD models again outperformed control-diffusion models ($R_\text{NERD}^2 = 0.869$; $R_\text{control-diffusion}^2 = 0.582$) in predicting human reward performance, and only NERD exhibited significant or trending cluster effects (\ref{tab:clusterInteractions}). Crucially, including clusters improved NERD models' predictive power dramatically (from $R^2 = 0.782$); while the addition of clusters also caused numerical improvement in goodness of fit for the control-diffusion models (from $R^2 = 0.321$), they still far underperformed NERD in predicting humans' achieved reward.

\section*{Discussion} % (1500 Words Maximum)

{\label{421360}}

% Discussion

Here, we developed the Noise Estimation through Reinforcement-based Diffusion (NERD) model to reveal and characterize humans' higher-order representations of the amount of variability expected in a first-order representation (about the world). We trained NERD models using fMRI data from three Decoded Neurofeedback (DecNef) tasks \citep{Cortese2021}, and compared their learned noise representations and capacity to capture human behavior to those of a control-diffusion model trained via backpropagation \citep{azimi2024diffusion}. By comparing NERD models' performance to that of control-diffusion models, we aimed to evaluate the hypothesis that human subjects solve DecNef through learning about variation in their first-order representations along task-relevent dimensions, that they do so using a reinforcement learning basis, and that NERD would thus be a powerful tool in extracting and characterizing higher-order representations about noise expectations in general.

We found that both NERD and control-diffusion models equivalently learned to denoise voxel patterns to match sample patterns of target pattern distributions. However, NERD better predicted human DecNef rewards
%($R_\text{NERD}^2 = 0.782$ vs. $R_\text{control-diffusion}^2 = 0.321$; Fig.~\ref{fig:nll}c,d) 
and exhibited more gradual denoising trajectories (Fig.~\ref{fig:rewardThruDenoising}). 
These findings provided initial evidence that NERD possesses improved capacity to comprehensively sample noise landscapes over the control-diffusion model. %Pattern similarity analyses also (Fig.~\ref{fig:RDMs}) confirmed NERD's richer exploration of voxel space, with trial-specific variability appearing minimal or absent in control-diffusion models. 

The critical analyses to test our hypotheses concerned directly examining NERD's learned noise distributions themselves, and evaluating whether they would capture significant individual variability in human DecNef performance. When we directly examined the noise distributions \( p(\mu, \sigma | t, x_t) \) (Fig.~\ref{fig:noiseTrajectories}), we observed heterogeneous, non-monotonic \( \mu \) and decreasing \( \sigma \) for NERD, which were quite unlike control-diffusion models' more monotonic patterns. We then used dimensionality reduction (PCA) to abstract from voxel space to something more akin to mental representatinoal space. This analysis showed higher effective dimensionality in NERD's noise trajectories: NERD's first principal component (PC1) exhibited a mean explained variance of [85.7\% $\pm$ 0.04, 78.45\% $\pm$ 0.12, 86.34\% $\pm$ 0.07], compared to control-diffusion's PC1 mean explained variance of [94.0\% $\pm$ 0.01, 89.6\% $\pm$ 0.3, 91.9\% $\pm$ 0.1]). NERD and control-diffusion also displayed different clusters (participants marked `similar' by NERD were often dissimilar for the control model and vice versa). Importantly, these clusters dramatically improved NERD's capacity to predict human-achieved rewards in the DecNef task.
%($R_\text{NERD}^2 = 0.869$ vs. $R_\text{control-diffusion}^2 = 0.782, 0.76, 0.73$; 
Thus, by training a denoising diffusion model to accomplish this denoising goal specifically using reinforcement learning approaches \citep{azimi2024diffusion}, we were able to discover and characterize higher-order expectations of neural representational noise and show that clusters of such learned noise distributions can explain individual differences in how well humans solve the DecNef task.

Our results support our hypothesis that humans may solve DecNef tasks by learning about and then minimizing the uncertainty in their first-order representations along the task-relevant dimension. Uncertainty reduction is a core cognitive function used in many other goal-directed behaviors \citep{gottlieb2013information,gottlieb2018towards, Cockburn2022Novelty, Gottlieb2012Attention, Jiwa2024Generating, Mobbs2018Foraging, Inglis2000Central}, so it is reasonable that the brain may co-opt this existing ability to solve the novel DecNef task. Further, this metacognitive capacity to monitor and then seek to minimize internal representational uncertainty may be particularly relevant for DecNef: Cortese and colleagues \citep{cortese2020unconscious} reported that participants with higher metacognitive sensitivity -- that is, higher introspective ability to track their task accuracy using confidence judgments -- showed superior performance in the reinforcement learning task embedded within a DecNef paradigm, selecting reward-maximizing actions at higher rates during the unconscious learning sessions.
%showed superior DecNef performance, particularly in the initial stages of learning to select optimal actions based on hidden brain states \citep{cortese2020unconscious}. 
In other statistical learning tasks, higher metacognitive sensitivity is associated with faster learning \citep{Hainguerlot2018Metacognitive}.
Our findings thus add to the growing literature on how humans solve DecNef, adding nuance to these and other previous explanations (such as the targeted neural plasticity model \citep{Shibata2021}). If humans indeed solve DecNef through a metacognitive process of learning and then minimizing first-order representational uncertainty, further exploration of individual variability in this capacity (Fig.~\ref{fig:noiseDistributions}d) could help optimize DecNef protocols and explain individual differences in learning ability \citep{taschereau2022real, Taschereau2018,cortese2022adaptive}. Ongoing work in our group seeks to apply our NERD approach to other DecNef studies \citep{Cortese2021} to seek general mechanisms of DecNef success or failure across neural pattern targets.

%Comparison to Existing Methods

From a methodological perspective, NERD differs in important ways from previous analytic approaches to characterizing neural noise, including state-of-the-art statistical methods like GLMsingle \citep{prince2022improving} and Generative Modeling of Signal and Noise (GSN) \citep{Kay2024Disentangling}. These methods estimate voxel noise with the goal of enhancing the reliability of BOLD signal measurements, but do not model how \textit{brains} learn about their \textit{own} uncertainty and so cannot be assumed to capture such metacognitive monitoring or higher-order representation formation. This interpretation is also bolstered by our finding that a control-diffusion model, trained with standard backpropagation rather than reinforcement learning, was substantively less able to capture human performance than NERD. Similarly, electrophysiological methods \citep{Pospisil2024Revisiting, Stringer2019HighDimensional, Williams2021Statistical} focus on neural noise estimation from a statistical measurement perspective, not higher-order metacognitive monitoring or learning processes. Alternative methods such as `The Algorithm Formerly Known as PRINCE' (TAFKAP) \citep{vanbergen2021tafkap, vanBergen2015Sensory} -- which rely on the probabilistic population coding framework supposing that neural population response represents the parameters of a probability distribution over environmental variables \citep{ma2006bayesian, ma2009population, meyniel2015confidence, walker2023studying} -- may appear closer to our target of monitoring uncertainty in first-order representations about the environment. However, TAFKAP in fact still estimates first-order representational uncertainty directly from the experimenter's viewpoint rather than capturing metacognitive or higher-order monitoring that the brain itself would undertake. NERD's strength lies in mimicking the human brain's learning about its own noise via reinforcement learning, offering a novel framework to study such higher-order representations about first-order noise. Future work could integrate TAFKAP or GSN with reinforcement learning to directly estimate first-order noise in a probabilistic, model-driven framework \citep{cunningham2014dimensionality, pang2016dimensionality, jazayeri2021interpreting}, including exploring different options for the structure of noise other than the standard Gaussian assumption we applied here.

%Limitations

We acknowledge that our study is limited in its generalizability because we focused on the cingulate and early visual cortex \citep{Cortese2021}: higher-order representations about uncertainty likely vary across regions in which the target first-order representation is presumably housed (e.g., ventral temporal cortex). Future work may wish to employ NERD or similar approaches to whole-brain DecNef data \citep{Cortese2021} data understand the homogeneity or heterogeneity of higher-order representations depending on their first-order targets. Additionally, we acknowledge that the Markov Decision Process and Gaussian noise assumptions we applied \citep{black2023training, sohl2015deep} may fail to capture the processes noise distributions actually present in the brain or the mental structures its ongoing activity represents. Future studies should therefore also should also explore the consequences of augmenting NERD with non-Gaussian noise assumptions or other advanced reinforcement learning training algorithms \citep{schulman2017ppo, williams1992simple}. 

We also assumed that voxel-level noise could be abstracted to `mental space' via dimensionality reduction (via PCA) \citep{Bishop2006, Schneider2023Learnable, Steinmetz2021Neuropixels, Stringer2019HighDimensional,Pearson1901,VazquezGarcia2024Review} to approximate first-order representational noise, especially along the task relevant dimension(s). The relationship between neural patterns and ``what they actually represent" in mental space remains an ongoing, active debate \citep{baker2022three,favela2023investigating, favela2025contextualizing, Machery2025Neural,Vilarroya2017Neural, mathis2024decoding}. As mentioned above, future studies may integrate direct estimation of first-order uncertainty into NERD, e.g. through methods such as TAFKAP \citep{vanbergen2021tafkap}, GSN \citep{Kay2024Disentangling}, or deep learning-based dimensionality reduction \citep{Bonnen2021Ventral, Cao2024Explanatory, cross2021using, DiCarlo2023ImageComputable, DiCarlo2022Recurrent, Finzi2022convolutional, Reimer2014ContextDependent, Zhuang2022Unsupervised,Yang2020Artificial, Zhuang2021Unsupervised}, to enable more direct first-order uncertainty estimation in `mental space' -- including addressing some of these methods' reliance on the theoretical commitments of probabilistic population coding theories \citep{ma2006bayesian, ma2009population, walker2023studying}.

%Conclusion

In summary, here we have shown that our Noise Estimation through Reinforcement-based Diffusion (NERD) model effectively captures higher-order expectations about noise that subjects can learn to exploit in DecNef. Our approach thus advances our understanding of how and why brains learn about their own uncertainty, with additional potential for optimizing DecNef procedures and more comprehensively exploring neural representations in general \citep{amano2016learning, Bao2020Map, Brouwer2011CrossOrientation, Chang2021Explaining, Goris2014Partitioning, Jehee2012Perceptual, Kamitani2005Decoding, kay2008identifying, Haynes2005Predicting, Serences2009Estimating, Smith2008Spatial}.
\newpage
\selectlanguage{english}
\FloatBarrier
\bibliographystyle{jneurosci}
\bibliography{references.bib%
}

@article{Vilarroya2017Neural,
	title        = {Neural Representation: A Survey-Based Analysis of the Notion},
	author       = {Vilarroya, Oscar},
	year         = 2017,
	journal      = {Frontiers in Psychology},
	volume       = 8,
	pages        = 1458,
	doi          = {10.3389/fpsyg.2017.01458},
	url          = {https://pubmed.ncbi.nlm.nih.gov/28900406/}
}

@article{Machery2025Neural,
  title={The concept of representation in the brain sciences: The current status and ways forward},
  author={Favela, Luis H and Machery, Edouard},
  journal={Mind \& Language},
  volume={40},
  number={2},
  pages={215--225},
  year={2025},
  publisher={Wiley Online Library}
}

@article{Cortese2021,
	title        = {The DecNef collection, fMRI data from closed-loop decoded neurofeedback experiments},
	author       = {Cortese, Aurelio and Tanaka, Saori C. and Amano, Kaoru and Koizumi, Ai and Lau, Hakwan and Sasaki, Yuka and Shibata, Kazuhisa and Taschereau-Dumouchel, Vincent and Watanabe, Takeo and Kawato, Mitsuo},
	year         = 2021,
	journal      = {Scientific Data},
	publisher    = {Nature Publishing Group},
	volume       = 8,
	number       = 1,
	pages        = 69,
	doi          = {10.1038/S41597-021-00845-7}
}

@article{Koizumi2016,
	title        = {Fear reduction without fear through reinforcement of neural activity that bypasses conscious exposure},
	author       = {Koizumi, Ai and Amano, Kaoru and Cortese, Aurelio and Lau, Hakwan and Kawato, Mitsuo},
	year         = 2016,
	journal      = {Nature Human Behaviour},
	doi          = {10.1038/s41562-016-0006}
}

@article{laconte2011decoding,
	title        = {Decoding fMRI brain states in real-time},
	author       = {LaConte, Stephen M},
	year         = 2011,
	journal      = {Neuroimage},
	publisher    = {Elsevier},
	volume       = 56,
	number       = 2,
	pages        = {440--454}
}

@article{watanabe2017advances,
	title        = {Advances in fMRI real-time neurofeedback},
	author       = {Watanabe, Takeo and Sasaki, Yuka and Shibata, Kazuhisa and Kawato, Mitsuo},
	year         = 2017,
	journal      = {Trends in cognitive sciences},
	publisher    = {Elsevier},
	volume       = 21,
	number       = 12,
	pages        = {997--1010}
}

@article{amano2016learning,
	title        = {Learning to associate orientation with color in early visual areas by associative decoded fMRI neurofeedback},
	author       = {Amano, Kaoru and Shibata, Kazuhisa and Kawato, Mitsuo and Sasaki, Yuka and Watanabe, Takeo},
	year         = 2016,
	journal      = {Current Biology},
	publisher    = {Elsevier},
	volume       = 26,
	number       = 14,
	pages        = {1861--1866}
}

@article{cortese2017decoded,
	title        = {Decoded fMRI neurofeedback can induce bidirectional confidence changes within single participants},
	author       = {Cortese, Aurelio and Amano, Kaoru and Koizumi, Ai and Lau, Hakwan and Kawato, Mitsuo},
	year         = 2017,
	journal      = {NeuroImage},
	publisher    = {Elsevier},
	volume       = 149,
	pages        = {323--337}
}

@article{taschereau2022real,
	title        = {Real-time functional MRI in the treatment of mental health disorders},
	author       = {Taschereau-Dumouchel, Vincent and Cushing, Cody A and Lau, Hakwan},
	year         = 2022,
	journal      = {Annual review of clinical psychology},
	publisher    = {Annual Reviews},
	volume       = 18,
	pages        = {125--154}
}

@article{Schonemann1966,
	title        = {A Generalized Solution of the Orthogonal Procrustes Problem},
	author       = {Sch{\"o}nemann, Peter H.},
	year         = 1966,
	journal      = {Psychometrika},
	volume       = 31,
	number       = 1,
	pages        = {1--10},
	doi          = {10.1007/BF02289451}
}

@article{Brown2019Understanding,
	title        = {Understanding the Higher-Order Approach to Consciousness},
	author       = {Brown, Richard and Lau, Hakwan and LeDoux, Joseph E.},
	year         = 2019,
	journal      = {Trends in Cognitive Sciences},
	volume       = 23,
	number       = 9,
	pages        = {754--768},
	doi          = {10.1016/j.tics.2019.06.009},
	url          = {https://pubmed.ncbi.nlm.nih.gov/31375408/}
}

@article{ho2020denoising,
	title        = {Denoising Diffusion Probabilistic Models},
	author       = {Ho, Jonathan and Jain, Ajay and Abbeel, Pieter},
	year         = 2020,
	journal      = {arXiv preprint arXiv:2006.11239}
}

@article{sohl2015deep,
	title        = {Deep Unsupervised Learning using Nonequilibrium Thermodynamics},
	author       = {Sohl-Dickstein, Jascha and Weiss, Eric and Maheswaranathan, Niru and Ganguli, Surya},
	year         = 2015,
	journal      = {arXiv preprint arXiv:1503.03585}
}

@article{kingma2021variational,
	title        = {Variational Diffusion Models},
	author       = {Kingma, Diederik P and Dhariwal, Prafulla},
	year         = 2021,
	journal      = {arXiv preprint arXiv:2107.00630}
}

@article{dhariwal2021diffusion,
	title        = {Diffusion Models Beat GANs on Image Synthesis},
	author       = {Dhariwal, Prafulla and Nichol, Alex},
	year         = 2021,
	journal      = {arXiv preprint arXiv:2105.05233}
}

@article{nichol2021improved,
	title        = {Improved Denoising Diffusion Probabilistic Models},
	author       = {Nichol, Alexander Q and Dhariwal, Prafulla},
	year         = 2021,
	journal      = {arXiv preprint arXiv:2102.09672}
}

@article{song2021scorebased,
	title        = {Score-Based Generative Modeling through Stochastic Differential Equations},
	author       = {Song, Yang and Sohl-Dickstein, Jascha and Kingma, Diederik P and Kumar, Abhishek and Ermon, Stefano and Poole, Ben},
	year         = 2021,
	journal      = {arXiv preprint arXiv:2011.13456}
}

@article{schulman2017ppo,
	title        = {Proximal Policy Optimization Algorithms},
	author       = {Schulman, John and Wolski, Filip and Dhariwal, Prafulla and Radford, Alec and Klimov, Oleg},
	year         = 2017,
	journal      = {arXiv preprint arXiv:1707.06347}
}

@article{black2023training,
	title        = {Training diffusion models with reinforcement learning},
	author       = {Black, Kevin and Janner, Michael and Du, Yilun and Kostrikov, Ilya and Levine, Sergey},
	year         = 2023,
	journal      = {arXiv preprint arXiv:2305.13301}
}

@article{williams1992simple,
	title        = {Simple statistical gradient-following algorithms for connectionist reinforcement learning},
	author       = {Williams, Ronald J},
	year         = 1992,
	journal      = {Machine learning},
	publisher    = {Springer},
	volume       = 8,
	number       = {3-4},
	pages        = {229--256}
}

@book{sutton2018reinforcement,
	title        = {Reinforcement Learning: An Introduction},
	author       = {Sutton, Richard S and Barto, Andrew G},
	year         = 2018,
	publisher    = {MIT press}
}

@article{shibata_neurofeedback2016,
	title        = {Differential Activation Patterns in the Same Brain Region Led to Opposite Emotional States},
	author       = {Shibata, Kazuhisa and Watanabe, Takeo and Kawato, Mitsuo and Sasaki, Yuka},
	year         = 2016,
	journal      = {PLOS Biology},
	doi          = {10.1371/journal.pbio.1002546}
}

@article{kay2008identifying,
	title        = {Identifying natural images from human brain activity},
	author       = {Kay, Kendrick N and Naselaris, Thomas and Prenger, Ryan J and Gallant, Jack L},
	year         = 2008,
	journal      = {Nature},
	publisher    = {Nature Publishing Group UK London},
	volume       = 452,
	number       = 7185,
	pages        = {352--355}
}

@article{Taschereau2018,
	title        = {Towards an unconscious neural reinforcement intervention for common fears},
	author       = {Vincent Taschereau-Dumouchel and Aurelio Cortese and Toshinori Chiba and J. D. Knotts and Mitsuo Kawato and Hakwan Lau},
	year         = 2018,
	journal      = {Proceedings of the National Academy of Sciences},
	volume       = 115,
	number       = 13,
	pages        = {3470--3475},
	doi          = {10.1073/pnas.1721572115}
}

@article{Jehee2012Perceptual,
	title        = {Perceptual learning selectively refines orientation representations in early visual cortex},
	author       = {Jehee, Janneke F. M. and Ling, Sam and Swisher, Jascha D. and van Bergen, Ruben S. and Tong, Frank},
	year         = 2012,
	journal      = {Journal of Neuroscience},
	volume       = 32,
	number       = 47,
	pages        = {16747--16753},
	doi          = {10.1523/JNEUROSCI.6112-11.2012},
	url          = {https://pubmed.ncbi.nlm.nih.gov/23175828/}
}

@article{azimi2024diffusion,
	title        = {Diffusion models and reinforcement learning: Novel pathways to modeling decoded fMRI neurofeedback},
	author       = {Azimi Azrari, Hojjat and Peters, Megan AK},
	year         = 2024,
	journal      = {Proceedings of the Cognitive Computational Neuroscience Meeting}
}

@article{baker2022three,
	title        = {Three aspects of representation in neuroscience},
	author       = {Baker, Ben and Lansdell, Benjamin and Kording, Konrad P},
	year         = 2022,
	journal      = {Trends in cognitive sciences},
	publisher    = {Elsevier},
	volume       = 26,
	number       = 11,
	pages        = {942--958}
}

@article{favela2023investigating,
	title        = {Investigating the concept of representation in the neural and psychological sciences},
	author       = {Favela, Luis H and Machery, Edouard},
	year         = 2023,
	journal      = {Frontiers in Psychology},
	publisher    = {Frontiers Media SA},
	volume       = 14,
	pages        = 1165622
}

@article{favela2025contextualizing,
	title        = {Contextualizing, eliminating, or glossing: What to do with unclear scientific concepts like representation},
	author       = {Favela, Luis H and Machery, Edouard},
	year         = 2025,
	journal      = {Mind \& Language},
	publisher    = {Wiley Online Library}
}

@article{tarr2002visual,
	title        = {Visual object recognition},
	author       = {Tarr, Michael J and Vuong, Quoc C},
	year         = 2002,
	journal      = {Steven’s handbook of experimental psychology},
	volume       = 1,
	pages        = {287--314}
}

@article{squire1991medial,
	title        = {The medial temporal lobe memory system},
	author       = {Squire, Larry R and Zola-Morgan, Stuart},
	year         = 1991,
	journal      = {Science},
	publisher    = {American Association for the Advancement of Science},
	volume       = 253,
	number       = 5026,
	pages        = {1380--1386}
}

@article{miller2001integrative,
	title        = {An integrative theory of prefrontal cortex function},
	author       = {Miller, Earl K and Cohen, Jonathan D},
	year         = 2001,
	journal      = {Annu. Rev. Neurosci.},
	publisher    = {Annual Reviews},
	volume       = 24,
	number       = {},
	pages        = {167--202}
}

@article{cleeremans2007conscious,
	title        = {Conscious access to first-order and higher-order representations},
	author       = {Cleeremans, Axel and Timmermans, Bert and Pasquali, Antoine},
	year         = 2007,
	journal      = {Trends Cogn. Sci.},
	publisher    = {Elsevier},
	volume       = 11,
	number       = 11,
	pages        = {465--472}
}

@article{fleming2020awareness,
	title        = {Awareness as inference in a higher-order state space},
	author       = {Fleming, Stephen M},
	year         = 2020,
	journal      = {Neuroscience of consciousness},
	publisher    = {Oxford University Press},
	volume       = 2020,
	number       = 1,
	pages        = {niz020}
}

@article{scikit-learn,
	title        = {Scikit-learn: Machine Learning in Python},
	author       = {Pedregosa, F. and Varoquaux, G. and Gramfort, A. and Michel, V. and Thirion, B. and Grisel, O. and Blondel, M. and Prettenhofer, P. and Weiss, R. and Dubourg, V. and Vanderplas, J. and Passos, A. and Cournapeau, D. and Brucher, M. and Perrot, M. and Duchesnay, É.},
	year         = 2011,
	journal      = {Journal of Machine Learning Research},
	volume       = 12,
	pages        = {2825--2830}
}

@article{lau2019consciousness,
	title        = {Consciousness, metacognition, \& perceptual reality monitoring},
	author       = {Lau, Hakwan},
	year         = 2019,
	journal      = {PsyArxiv}
}

@incollection{michelperceptual,
	title        = {The {Perceptual} {Reality} {Monitoring} {Theory} (1st edition)},
	author       = {Michel, Matthias},
	year         = 2024,
	booktitle    = {Scientific {Theories} of {Consciousness}: {The} {Grand} {Tour}},
	publisher    = {Cambridge University Press},
	editor       = {Herzog, Michael and Schurger, Aaron and Doerig, Adrien}
}

@article{winter2022variance,
	title        = {Variance misperception under skewed empirical noise statistics explains overconfidence in the visual periphery},
	author       = {Winter, Charles J and Peters, Megan AK},
	year         = 2022,
	journal      = {Attention, Perception, \& Psychophysics},
	publisher    = {Springer},
	volume       = 84,
	number       = 1,
	pages        = {161--178}
}

@article{peters2025introspective,
	title        = {Introspective psychophysics for the study of subjective experience},
	author       = {Peters, Megan AK},
	year         = 2025,
	journal      = {Cerebral Cortex},
	publisher    = {Oxford University Press},
	volume       = 35,
	number       = 1,
	pages        = {49--57}
}

@article{peters2022towards,
	title        = {Towards characterizing the canonical computations generating phenomenal experience},
	author       = {Peters, Megan A. K.},
	year         = 2022,
	journal      = {Neurosci. Biobehav. Rev.},
	publisher    = {Elsevier},
	volume       = 142,
	pages        = 104903
}

@article{meyniel2015confidence,
	title        = {Confidence as a metacognitive source of learning speed},
	author       = {Meyniel, Florent and Sigman, Mariano and Mainen, Zachary F},
	year         = 2015,
	journal      = {Nat. Rev. Neurosci.},
	publisher    = {Nature Publishing Group},
	volume       = 16,
	number       = 12,
	pages        = {721--729}
}

@article{guggenmos2016mesolimbic,
	title        = {Mesolimbic confidence signals guide perceptual learning in the absence of external feedback},
	author       = {Guggenmos, Matthias and Wilbertz, Gregor and Hebart, Martin N and Sterzer, Philipp},
	year         = 2016,
	journal      = {eLife},
	publisher    = {eLife Sciences Publications},
	volume       = 5,
	pages        = {e13388}
}

@article{ma2006bayesian,
	title        = {Bayesian inference with probabilistic population codes},
	author       = {Ma, Wei Ji and Beck, Jeffrey M and Latham, Peter E and Pouget, Alexandre},
	year         = 2006,
	journal      = {Nat. Neurosci.},
	publisher    = {Nature Publishing Group},
	volume       = 9,
	number       = 11,
	pages        = {1432--1438}
}

@incollection{ma2009population,
	title        = {Population codes, correlations, and coding},
	author       = {Ma, Wei Ji and Pouget, Alexandre},
	year         = 2009,
	booktitle    = {The Cognitive Neurosciences},
	publisher    = {MIT Press},
	pages        = {135--144},
	editor       = {Gazzaniga, Michael S},
	edition      = {4th}
}

@article{vanbergen2021tafkap,
	title        = {TAFKAP: An improved method for probabilistic decoding of cortical activity},
	author       = {van Bergen, Ruben S and Jehee, Janneke F M},
	year         = 2021,
	journal      = {bioRxiv},
	publisher    = {Cold Spring Harbor Laboratory},
	doi          = {10.1101/2021.03.04.433946}
}

@article{prince2022improving,
	title        = {Improving the accuracy of single-trial fMRI response estimates using GLMsingle},
	author       = {Prince, Jacob S and Charest, Ian and Kurzawski, Jan W and Pyles, John A and Tarr, Michael J and Kay, Kendrick N},
	year         = 2022,
	journal      = {Elife},
	publisher    = {eLife Sciences Publications Limited},
	volume       = 11,
	pages        = {e77599}
}

@article{cunningham2014dimensionality,
	title        = {Dimensionality reduction for large-scale neural recordings},
	author       = {Cunningham, John P and Yu, Byron M},
	year         = 2014,
	journal      = {Nature neuroscience},
	publisher    = {Nature Publishing Group US New York},
	volume       = 17,
	number       = 11,
	pages        = {1500--1509}
}

@article{jazayeri2021interpreting,
	title        = {Interpreting neural computations by examining intrinsic and embedding dimensionality of neural activity},
	author       = {Jazayeri, Mehrdad and Ostojic, Srdjan},
	year         = 2021,
	journal      = {Current opinion in neurobiology},
	publisher    = {Elsevier},
	volume       = 70,
	pages        = {113--120}
}

@article{pang2016dimensionality,
	title        = {Dimensionality reduction in neuroscience},
	author       = {Pang, Rich and Lansdell, Benjamin J and Fairhall, Adrienne L},
	year         = 2016,
	journal      = {Current Biology},
	publisher    = {Elsevier},
	volume       = 26,
	number       = 14,
	pages        = {R656--R660}
}

@article{walker2023studying,
	title        = {Studying the neural representations of uncertainty},
	author       = {Walker, Edgar Y and Pohl, Stephan and Denison, Rachel N and Barack, David L and Lee, Jennifer and Block, Ned and Ma, Wei Ji and Meyniel, Florent},
	year         = 2023,
	journal      = {Nat. Neurosci.},
	publisher    = {Nature Publishing Group},
	volume       = 26,
	number       = 11,
	pages        = {1857--1867},
	doi          = {10.1038/s41593-023-01444-y}
}

@article{shekhar2024humans,
  title={How do you get there from here? Mental representations of route descriptions},
  author={Schneider, Laura F and Taylor, Holly A},
  journal={Applied Cognitive Psychology: The Official Journal of the Society for Applied Research in Memory and Cognition},
  volume={13},
  number={5},
  pages={415--441},
  year={1999},
  publisher={Wiley Online Library}
}

@article{Froemer2021Response,
	title        = {Response-based outcome predictions and confidence regulate feedback processing and learning},
	author       = {Frömer, Romy and Nassar, Matthew R. and Bruckner, Rasmus and Stürmer, Birgit and Sommer, Werner and Yeung, Nick},
	year         = 2021,
	journal      = {eLife},
	volume       = 10,
	pages        = {e62825},
	doi          = {10.7554/eLife.62825},
	url          = {https://elifesciences.org/articles/62825}
}

@article{Hainguerlot2018Metacognitive,
	title        = {Metacognitive ability predicts learning cue-stimulus associations in the absence of external feedback},
	author       = {Hainguerlot, Marine and Vergnaud, Jean-Christophe and de Gardelle, Vincent},
	year         = 2018,
	journal      = {Scientific Reports},
	volume       = 8,
	number       = 1,
	pages        = 5602,
	doi          = {10.1038/s41598-018-23936-9},
	url          = {https://www.nature.com/articles/s41598-018-23936-9}
}

@article{Boundy-Singer2023Confidence,
	title        = {Confidence reflects a noisy decision reliability estimate},
	author       = {Boundy-Singer, Zoe M. and Ziemba, Corey M. and Goris, Robbe L. T.},
	year         = 2023,
	journal      = {Nature Human Behaviour},
	volume       = 7,
	number       = 1,
	pages        = {142--154},
	doi          = {10.1038/s41562-022-01464-x},
	url          = {https://www.nature.com/articles/s41562-022-01464-x}
}

@article{Mamassian2024CNCB,
	title        = {The Confidence-Noise Confidence-Boost (CNCB) model of confidence rating data},
	author       = {Mamassian, Pascal and de Gardelle, Vincent},
	year         = 2024,
	journal      = {bioRxiv},
	doi          = {10.1101/2024.09.04.611165},
	url          = {https://www.biorxiv.org/content/10.1101/2024.09.04.611165v2}
}

@article{mamassian2018confidence,
	title        = {Confidence forced-choice and other metaperceptual tasks},
	author       = {Mamassian, Pascal},
	year         = 2018,
	journal      = {Perception},
	publisher    = {SAGE Publications},
	volume       = 47,
	number       = {10-11},
	pages        = {1023--1035},
	doi          = {10.1177/0301006618790116}
}

@article{mamassian2022modeling,
	title        = {Modeling perceptual confidence and the confidence forced-choice paradigm},
	author       = {Mamassian, Pascal and de Gardelle, Vincent},
	year         = 2022,
	journal      = {Psychol. Rev.},
	publisher    = {American Psychological Association},
	volume       = 129,
	number       = 5,
	pages        = {976--998}
}

@article{Gottlieb2012Attention,
	title        = {Attention, learning, and the value of information},
	author       = {Gottlieb, Jacqueline},
	year         = 2012,
	journal      = {Neuron},
	volume       = 76,
	number       = 2,
	pages        = {281--295},
	doi          = {10.1016/j.neuron.2012.09.034},
	url          = {https://pubmed.ncbi.nlm.nih.gov/23083732/}
}

@article{Jiwa2024Generating,
	title        = {Generating Saccades for Reducing Uncertainty: Cognitive and Sensorimotor Trade-Offs},
	author       = {Jiwa, Matthew and Rothkopf, Constantin and Gottlieb, Jacqueline},
	year         = 2024,
	journal      = {Journal of Vision},
	volume       = 24,
	number       = 908,
	doi          = {10.1167/jov.24.10.908},
	url          = {https://doi.org/10.1167/jov.24.10.908}
}

@article{peters2017perceptual,
	title        = {Perceptual confidence neglects decision-incongruent evidence in the brain},
	author       = {Peters, Megan AK and Thesen, Thomas and Ko, Yoshiaki D and Maniscalco, Brian and Carlson, Chad and Davidson, Matt and Doyle, Werner and Kuzniecky, Ruben and Devinsky, Orrin and Halgren, Eric and others},
	year         = 2017,
	journal      = {Nature human behaviour},
	publisher    = {Nature Publishing Group UK London},
	volume       = 1,
	number       = 7,
	pages        = {0139}
}

@article{gottlieb2018towards,
	title        = {Towards a neuroscience of active sampling and curiosity},
	author       = {Gottlieb, Jacqueline and Oudeyer, Pierre-Yves},
	year         = 2018,
	journal      = {Nature Reviews Neuroscience},
	publisher    = {Nature Publishing Group},
	volume       = 19,
	number       = 12,
	pages        = {758--770},
	doi          = {10.1038/s41583-018-0078-0}
}

@article{gottlieb2013information,
	title        = {Information-seeking, curiosity, and attention: computational and neural mechanisms},
	author       = {Gottlieb, Jacqueline and Oudeyer, Pierre-Yves and Lopes, Manuel and Baranes, Adrien},
	year         = 2013,
	journal      = {Trends in Cognitive Sciences},
	publisher    = {Elsevier},
	volume       = 17,
	number       = 11,
	pages        = {585--593},
	doi          = {10.1016/j.tics.2013.09.001}
}

@article{Meyniel2017Brain,
	title        = {Brain networks for confidence weighting and hierarchical inference during probabilistic learning},
	author       = {Meyniel, Florent and Dehaene, Stanislas},
	year         = 2017,
	journal      = {Proceedings of the National Academy of Sciences},
	volume       = 114,
	number       = 19,
	pages        = {E3859--E3868},
	doi          = {10.1073/pnas.1615773114},
	url          = {https://www.pnas.org/doi/10.1073/pnas.1615773114}
}

@article{Guggenmos2022Reverse,
	title        = {Reverse engineering of metacognition},
	author       = {Guggenmos, Matthias},
	year         = 2022,
	journal      = {eLife},
	volume       = 11,
	pages        = {e75420},
	doi          = {10.7554/eLife.75420},
	url          = {https://elifesciences.org/articles/75420}
}

@article{Meyniel2015Sense,
	title        = {The Sense of Confidence during Probabilistic Learning: A Normative Account},
	author       = {Meyniel, Florent and Schlunegger, Daniel and Dehaene, Stanislas},
	year         = 2015,
	journal      = {PLoS Computational Biology},
	volume       = 11,
	number       = 6,
	pages        = {e1004305},
	doi          = {10.1371/journal.pcbi.1004305},
	url          = {https://journals.plos.org/ploscompbiol/article?id=10.1371/journal.pcbi.1004305}
}

@article{cross2021using,
	title        = {Using deep reinforcement learning to reveal how the brain encodes abstract state-space representations in high-dimensional environments},
	author       = {Cross, L and Cockburn, J and Yue, Y and O'Doherty, JP},
	year         = 2021,
	journal      = {Neuron},
	publisher    = {Elsevier},
	volume       = 109,
	number       = 4,
	pages        = {724--738.e7},
	doi          = {10.1016/j.neuron.2020.11.021},
	epub         = {2020 Dec 15},
	pmid         = 33326755,
	pmcid        = {PMC7897245}
}

@article{lebel2021voxelwise,
	title        = {Voxelwise encoding models show that cerebellar language representations are highly conceptual},
	author       = {LeBel, A and Jain, S and Huth, A G},
	year         = 2021,
	journal      = {Journal of Neuroscience},
	publisher    = {Society for Neuroscience},
	volume       = 41,
	number       = 50,
	pages        = {10341--10355}
}

@article{nishimoto2011reconstructing,
	title        = {Reconstructing visual experiences from brain activity evoked by natural movies},
	author       = {Nishimoto, S and Vu, A T and Naselaris, T and Benjamini, Y and Yu, B and Gallant, J L},
	year         = 2011,
	journal      = {Current Biology},
	publisher    = {Elsevier},
	volume       = 21,
	number       = 19,
	pages        = {1641--1646}
}

@article{nunez2019voxelwise,
	title        = {Voxelwise encoding models with non-spherical multivariate normal priors},
	author       = {Nunez-Elizalde, A O and Huth, A G and Gallant, J L},
	year         = 2019,
	journal      = {NeuroImage},
	publisher    = {Elsevier},
	volume       = 197,
	pages        = {482--492}
}

@article{rumelhart1986learning,
	title        = {Learning representations by back-propagating errors},
	author       = {Rumelhart, David E and Hinton, Geoffrey E and Williams, Ronald J},
	year         = 1986,
	journal      = {Nature},
	publisher    = {Nature Publishing Group},
	volume       = 323,
	number       = 6088,
	pages        = {533--536},
	doi          = {10.1038/323533a0}
}

@article{jonsdottir2013levy,
	title        = {Lévy-based Modelling in Brain Imaging},
	author       = {Jonsdottir, Kristjana Yr and Rønn-Nielsen, Anders and Mouridsen, Kim and Jensen, Eva B. Vedel},
	year         = 2013,
	journal      = {Scandinavian Journal of Statistics},
	publisher    = {Wiley},
	doi          = {10.1002/SJOS.12000}
}

@article{mausfeld2012cognitive,
	title        = {On Some Unwarranted Tacit Assumptions in Cognitive Neuroscience},
	author       = {Mausfeld, Rainer},
	year         = 2012,
	journal      = {Frontiers in Psychology},
	publisher    = {Frontiers},
	doi          = {10.3389/FPSYG.2012.00067}
}

@article{parker2022assumptions,
	title        = {Assumptions of Twentieth-Century Neuroscience: Reductionist and Computational Paradigms},
	author       = {Parker, David},
	year         = 2022,
	journal      = {Interdisciplinary Science Reviews},
	publisher    = {Taylor \& Francis},
	doi          = {10.1080/03080188.2022.2149736}
}

@article{dewit2019information,
	title        = {Is Information Theory, or the Assumptions That Surround It, Holding Back Neuroscience?},
	author       = {de-Wit, Lee and Ekroll, Vebjørn and Schwarzkopf, Dietrich Samuel and Wagemans, Johan},
	year         = 2019,
	journal      = {Behavioral and Brain Sciences},
	publisher    = {Cambridge University Press},
	doi          = {10.1017/S0140525X19001250}
}

@article{hancock1996face,
	title        = {Face processing: Human perception and principal components analysis},
	author       = {Hancock, Peter JB and Burton, A Mike and Bruce, Vicki},
	year         = 1996,
	journal      = {Memory \& cognition},
	publisher    = {Springer},
	volume       = 24,
	pages        = {26--40}
}

@article{rhodes2015distinct,
	title        = {How distinct is the coding of face identity and expression? Evidence for some common dimensions in face space},
	author       = {Rhodes, Gillian and Pond, Stephen and Burton, Nichola and Kloth, Nadine and Jeffery, Linda and Bell, Jason and Ewing, Louise and Calder, Andrew J and Palermo, Romina},
	year         = 2015,
	journal      = {Cognition},
	publisher    = {Elsevier},
	volume       = 142,
	pages        = {123--137}
}

@article{Pearson1901,
	title        = {On lines and planes of closest fit to systems of points in space},
	author       = {Karl Pearson},
	year         = 1901,
	journal      = {The London, Edinburgh, and Dublin Philosophical Magazine and Journal of Science},
	volume       = 2,
	number       = 11,
	pages        = {559--572},
	doi          = {10.1080/14786440109462720}
}

@book{Bishop2006,
	title        = {Pattern Recognition and Machine Learning},
	author       = {Christopher M. Bishop},
	year         = 2006,
	publisher    = {Springer},
	isbn         = {978-0-387-31073-2}
}

@article{Kay2024Disentangling,
	title        = {Disentangling signal and noise in neural responses through generative modeling},
	author       = {Kay, Kendrick N and Prince, Jacob S and Gebhart, Thomas and Tuckute, Greta and Zhou, Jingyang and Naselaris, Thomas and Schutt, Heiko},
	year         = 2024,
	journal      = {bioRxiv},
	doi          = {10.1101/2024.04.22.590510},
	url          = {https://pubmed.ncbi.nlm.nih.gov/38712051/}
}

@article{Yamins2014Performance,
	title        = {Performance-optimized hierarchical models predict neural responses in higher visual cortex},
	author       = {Yamins, Daniel L. K. and Hong, Ha and Cadieu, Charles F. and Solomon, Ethan A. and Seibert, Darren and DiCarlo, James J.},
	year         = 2014,
	journal      = {Proceedings of the National Academy of Sciences},
	volume       = 111,
	number       = 23,
	pages        = {8619--8624},
	doi          = {10.1073/pnas.1403112111},
	url          = {https://www.pnas.org/doi/10.1073/pnas.1403112111}
}

@article{Pospisil2024Revisiting,
	title        = {Revisiting the high-dimensional geometry of population responses in visual cortex},
	author       = {Pospisil, Dean A. and Pillow, Jonathan W.},
	year         = 2024,
	journal      = {bioRxiv},
	doi          = {10.1101/2024.02.16.580726},
	url          = {https://doi.org/10.1101/2024.02.16.580726}
}

@article{Williams2021Statistical,
	title        = {Statistical neuroscience in the single trial limit},
	author       = {Williams, Alex H. and Linderman, Scott W.},
	year         = 2021,
	journal      = {Current Opinion in Neurobiology},
	volume       = 70,
	pages        = {193--205},
	doi          = {10.1016/j.conb.2021.10.008},
	url          = {https://pubmed.ncbi.nlm.nih.gov/34861596/}
}

@article{vanBergen2015Sensory,
	title        = {Sensory uncertainty decoded from visual cortex predicts behavior},
	author       = {van Bergen, Ruben S. and Ma, Wei Ji and Pratte, Michael S. and Jehee, Janneke F. M.},
	year         = 2015,
	journal      = {Nature Neuroscience},
	volume       = 18,
	pages        = {1728--1730},
	doi          = {10.1038/nn.4150},
	url          = {https://www.nature.com/articles/nn.4150}
}

@article{Kamitani2005Decoding,
	title        = {Decoding the visual and subjective contents of the human brain},
	author       = {Kamitani, Yukiyasu and Tong, Frank},
	year         = 2005,
	journal      = {Nature Neuroscience},
	volume       = 8,
	number       = 5,
	pages        = {679--685},
	doi          = {10.1038/nn1444},
	url          = {https://www.nature.com/articles/nn1444}
}

@article{Haynes2005Predicting,
	title        = {Predicting the orientation of invisible stimuli from activity in human primary visual cortex},
	author       = {Haynes, John-Dylan and Rees, Geraint},
	year         = 2005,
	journal      = {Nature Neuroscience},
	volume       = 8,
	number       = 5,
	pages        = {686--691},
	doi          = {10.1038/nn1445},
	url          = {https://www.nature.com/articles/nn1445}
}

@article{Brouwer2011CrossOrientation,
	title        = {Cross-orientation suppression in human visual cortex},
	author       = {Brouwer, Gijs Joost and Heeger, David J.},
	year         = 2011,
	journal      = {Journal of Neurophysiology},
	volume       = 106,
	number       = 5,
	pages        = {2108--2119},
	doi          = {10.1152/jn.00540.2011},
	url          = {https://pubmed.ncbi.nlm.nih.gov/21775720/}
}

@article{Serences2009Estimating,
	title        = {Estimating the influence of attention on population codes in human visual cortex using voxel-based tuning functions},
	author       = {Serences, John T. and Saproo, Sameer and Scolari, Miranda and Ho, Tiffany and Muftuler, Levent T.},
	year         = 2009,
	journal      = {NeuroImage},
	volume       = 44,
	number       = 1,
	pages        = {223--231},
	doi          = {10.1016/j.neuroimage.2008.07.043},
	url          = {https://pubmed.ncbi.nlm.nih.gov/18793734/}
}

@article{Smith2008Spatial,
	title        = {Spatial and temporal scales of neuronal correlation in primary visual cortex},
	author       = {Smith, Matthew A. and Kohn, Adam},
	year         = 2008,
	journal      = {Journal of Neuroscience},
	volume       = 28,
	number       = 48,
	pages        = {12591--12603},
	doi          = {10.1523/JNEUROSCI.2929-08.2008},
	url          = {https://www.jneurosci.org/content/28/48/12591}
}

@article{Goris2014Partitioning,
	title        = {Partitioning neuronal variability},
	author       = {Goris, Robbe L.T. and Movshon, J. Anthony and Simoncelli, Eero P.},
	year         = 2014,
	journal      = {Nature Neuroscience},
	volume       = 17,
	number       = 6,
	pages        = {858--865},
	doi          = {10.1038/nn.3711},
	url          = {https://www.nature.com/articles/nn.3711}
}

@article{Chang2021Explaining,
	title        = {Explaining face representation in the primate brain using different computational models},
	author       = {Chang, Le and Egger, Bernhard and Vetter, Thomas and Tsao, Doris Y.},
	year         = 2021,
	journal      = {Current Biology},
	volume       = 31,
	number       = 14,
	pages        = {2940--2949.e4},
	doi          = {10.1016/j.cub.2021.05.064},
	url          = {https://www.sciencedirect.com/science/article/pii/S0960982221005273}
}

@article{Bao2020Map,
	title        = {A map of object space in primate inferotemporal cortex},
	author       = {Bao, Pinglei and She, Liang and McGill, Mason and Tsao, Doris Y.},
	year         = 2020,
	journal      = {Nature},
	volume       = 583,
	number       = 7814,
	pages        = {103--108},
	doi          = {10.1038/s41586-020-2350-5},
	url          = {https://pubmed.ncbi.nlm.nih.gov/32494012/}
}

@article{Cockburn2022Novelty,
	title        = {Novelty and uncertainty regulate the balance between exploration and exploitation through distinct mechanisms in the human brain},
	author       = {Cockburn, Jeffrey and Man, Vincent and Cunningham, William A. and O'Doherty, John P.},
	year         = 2022,
	journal      = {Neuron},
	volume       = 110,
	number       = 16,
	pages        = {2691--2702},
	doi          = {10.1016/j.neuron.2022.05.025},
	url          = {https://pubmed.ncbi.nlm.nih.gov/35809575/}
}

@article{Mobbs2018Foraging,
	title        = {Foraging for foundations in decision neuroscience: insights from ethology},
	author       = {Mobbs, Dean and Trimmer, Pete C. and Blumstein, Daniel T. and Dayan, Peter},
	year         = 2018,
	journal      = {Nature Reviews Neuroscience},
	volume       = 19,
	number       = 7,
	pages        = {419--427},
	doi          = {10.1038/s41583-018-0010-7},
	url          = {https://www.nature.com/articles/s41583-018-0010-7}
}

@article{Inglis2000Central,
	title        = {The Central Role of Uncertainty Reduction in Determining Behaviour},
	author       = {Inglis, Ian},
	year         = 2000,
	journal      = {Behaviour},
	volume       = 137,
	number       = 12,
	pages        = {1567--1599},
	doi          = {10.1163/156853900502727},
	url          = {https://brill.com/view/journals/beh/137/12/article-p1567_1.xml}
}

@article{Yang2020Artificial,
	title        = {Artificial Neural Networks for Neuroscientists: A Primer},
	author       = {Yang, Guangyu Robert and Wang, Xiao-Jing},
	year         = 2020,
	journal      = {Neuron},
	volume       = 107,
	number       = 6,
	pages        = {1048--1070},
	doi          = {10.1016/j.neuron.2020.09.005},
	url          = {https://doi.org/10.1016/j.neuron.2020.09.005}
}

@article{Cao2024Explanatory,
	title        = {Explanatory models in neuroscience, Part 1: Taking mechanistic abstraction seriously},
	author       = {Cao, Rosa and Yamins, Daniel},
	year         = 2024,
	journal      = {Cognitive Systems Research},
	volume       = 74,
	pages        = 101244,
	doi          = {10.1016/j.cogsys.2024.101244},
	url          = {https://www.sciencedirect.com/science/article/abs/pii/S138904172400038X}
}

@article{DiCarlo2023ImageComputable,
	title        = {Let's move forward: Image-computable models and a common model evaluation scheme are prerequisites for a scientific understanding of human vision},
	author       = {DiCarlo, James J. and Yamins, Daniel L. K. and Ferguson, Michael E. and Fedorenko, Evelina and Bethge, Matthias and Bonnen, Tyler and Schrimpf, Martin},
	year         = 2023,
	journal      = {Behavioral and Brain Sciences},
	volume       = 46,
	pages        = {e390},
	doi          = {10.1017/S0140525X23001607},
	url          = {https://www.cambridge.org/core/journals/behavioral-and-brain-sciences/article/lets-move-forward-imagecomputable-models-and-a-common-model-evaluation-scheme-are-prerequisites-for-a-scientific-understanding-of-human-vision/F2302912C8652DC2582F0E159C1BB6AB}
}

@article{DiCarlo2022Recurrent,
	title        = {Recurrent Connections in the Primate Ventral Visual Stream Mediate a Trade-Off Between Task Performance and Network Size During Core Object Recognition},
	author       = {DiCarlo, James J. and Yamins, Daniel L.K. and Ferguson, Michael E. and Fedorenko, Evelina and Bethge, Matthias and others},
	year         = 2022,
	journal      = {Neural Computation},
	volume       = 34,
	number       = 8,
	pages        = {1652--1676},
	doi          = {10.1162/neco_a_01506},
	url          = {https://direct.mit.edu/neco/article/34/8/1652/111780/Recurrent-Connections-in-the-Primate-Ventral}
}

@article{Zhuang2021Unsupervised,
	title        = {Unsupervised neural network models of the ventral visual stream},
	author       = {Zhuang, Chengxu and Yan, Siming and Nayebi, Aran and Schrimpf, Martin and Frank, Michael C. and DiCarlo, James J. and Yamins, Daniel L.K.},
	year         = 2021,
	journal      = {Proceedings of the National Academy of Sciences},
	volume       = 118,
	number       = 3,
	pages        = {e2014196118},
	doi          = {10.1073/pnas.2014196118},
	url          = {https://www.pnas.org/doi/abs/10.1073/pnas.2014196118}
}

@article{Bonnen2021Ventral,
	title        = {When the ventral visual stream is not enough: A deep learning account of medial temporal lobe involvement in perception},
	author       = {Bonnen, Tyler and Yamins, Daniel L.K. and Wagner, Anthony D.},
	year         = 2021,
	journal      = {Neuron},
	volume       = 109,
	number       = 17,
	pages        = {2755--2766.e6},
	doi          = {10.1016/j.neuron.2021.06.018},
	url          = {https://www.cell.com/neuron/fulltext/S0896-6273(21)00459-1}
}

@article{Finzi2022convolutional,
	title        = {Do deep convolutional neural networks accurately model representations beyond the ventral stream?},
	author       = {Finzi, Dawn and Yamins, Daniel L.K. and Kay, Kendrick N and Grill-Spector, Kalanit},
	year         = 2022,
	journal      = {Proceedings of the Cognitive Computational Neuroscience Conference},
	url          = {https://www.dawnfinzi.com/publication/ccn/ccn.pdf}
}

@article{Reimer2014ContextDependent,
	title        = {Context-dependent computation by recurrent dynamics in prefrontal cortex},
	author       = {Reimer, Michael L. and Mante, Mark M. and Minxha, Michael D. and {others}},
	year         = 2014,
	journal      = {Nature},
	volume       = 507,
	number       = 7493,
	pages        = {131--134},
	doi          = {10.1038/nature12742},
	url          = {https://www.nature.com/articles/nature12742}
}

@inproceedings{Zhuang2022Unsupervised,
	title        = {How Well Do Unsupervised Learning Algorithms Model Human Real-time and Life-long Learning?},
	author       = {Zhuang, Chengxu and Xiang, Ziyu and Bai, Yoon and Jia, Xiaoxuan and Turk-Browne, Nicholas and Norman, Kenneth and DiCarlo, James J. and Yamins, Daniel L.K.},
	year         = 2022,
	booktitle    = {Proceedings of the 36th Conference on Neural Information Processing Systems (NeurIPS 2022)},
	url          = {https://proceedings.neurips.cc/paper_files/paper/2022/hash/8dfc3a2720a4112243a285b98e0d4415-Abstract-Datasets_and_Benchmarks.html}
}

@article{Steinmetz2021Neuropixels,
	title        = {Neuropixels 2.0: A miniaturized high-density probe for stable, long-term brain recordings},
	author       = {Steinmetz, Nicholas A and Aydin, Cagatay and Lebedeva, Anna and Okun, Michael and Pachitariu, Marius and Bauza, Marius and Beau, Maxime and Bhagat, Jai and Böhm, Claudia and Broux, Martijn and Chen, Susu and Colonell, Jennifer and Gardner, Richard J and Karsh, Bill and Kloosterman, Fabian and Kostadinov, Dimitar and Mora-Lopez, Carolina and O'Callaghan, John and Park, Junchol and Putzeys, Jan and Sauerbrei, Britton and van Daal, Rik J J and Vollan, Abraham Z and Wang, Shiwei and Welkenhuysen, Marleen and Ye, Zhiwen and Dudman, Joshua T and Dutta, Barundeb and Hantman, Adam W and Harris, Kenneth D and Lee, Albert K and Moser, Edvard I and O'Keefe, John and Renart, Alfonso and Svoboda, Karel and Häusser, Michael and Haesler, Sebastian and Carandini, Matteo and Harris, Timothy D},
	year         = 2021,
	journal      = {Science},
	volume       = 372,
	number       = 6539,
	pages        = {eabf4588},
	doi          = {10.1126/science.abf4588},
	url          = {https://www.science.org/doi/abs/10.1126/science.abf4588}
}

@article{Stringer2019HighDimensional,
	title        = {High-dimensional geometry of population responses in visual cortex},
	author       = {Stringer, Carsen and Pachitariu, Marius and Steinmetz, Nicholas and Carandini, Matteo and Harris, Kenneth D.},
	year         = 2019,
	journal      = {Nature},
	volume       = 571,
	number       = 7765,
	pages        = {361--365},
	doi          = {10.1038/s41586-019-1346-5},
	url          = {https://www.nature.com/articles/s41586-019-1346-5}
}

@article{Schneider2023Learnable,
	title        = {Learnable latent embeddings for joint behavioural and neural analysis},
	author       = {Schneider, Steffen and Lee, Jin Hwa and Mathis, Mackenzie Weygandt},
	year         = 2023,
	journal      = {Nature},
	volume       = 617,
	pages        = {360--368},
	doi          = {10.1038/s41586-023-06031-6},
	url          = {https://www.nature.com/articles/s41586-023-06031-6}
}

@article{VazquezGarcia2024Review,
	title        = {A Review of Latent Representation Models in Neuroimaging},
	author       = {Vázquez-García, C. and Martínez-Murcia, F.J. and Segovia Román, F. and Górriz, Juan M.},
	year         = 2024,
	journal      = {arXiv preprint arXiv:2412.19844},
	url          = {https://arxiv.org/abs/2412.19844}
}

@article{kianishadlen2009,
	title        = {Representation of confidence associated with a decision by neurons in the parietal cortex},
	author       = {Kiani, Roozbeh and Shadlen, Michael N.},
	year         = 2009,
	journal      = {Science}
}

@article{odegaard2018pnas,
	title        = {Superior colliculus neuronal ensemble activity signals optimal rather than subjective confidence},
	author       = {Odegaard, Brian and Grimaldi, Piercesare and Hah Cho, Seong and Peters, Megan A. K. and Lau, Hakwan and Basso, Michele A.},
	year         = 2018,
	journal      = {Proceedings of the National Academy of Sciences},
	volume       = 115,
	number       = 7,
	pages        = {E1588-E1597},
	doi          = {10.1073/pnas.1711628115},
	url          = {https://www.pnas.org/doi/abs/10.1073/pnas.1711628115}
}

@article{Bang2018Distinct,
	title        = {Distinct encoding of decision confidence in human medial prefrontal cortex},
	author       = {Bang, Dan and Fleming, Stephen M.},
	year         = 2018,
	journal      = {Proceedings of the National Academy of Sciences},
	volume       = 115,
	number       = 23,
	pages        = {6082--6087},
	doi          = {10.1073/pnas.1800795115},
	url          = {https://doi.org/10.1073/pnas.1800795115}
}

@book{Knill1996,
	title        = {Perception as Bayesian Inference},
	year         = 1996,
	publisher    = {Cambridge University Press},
	isbn         = 9780521461092,
	url          = {https://www.cambridge.org/core/books/perception-as-bayesian-inference/0442F577F5E4CD874FA6819978574C8F},
	editor       = {David C. Knill and Whitman Richards}
}

@article{Shibata2021,
  title={Toward a comprehensive understanding of the neural mechanisms of decoded neurofeedback},
  author={Shibata, Kazuhisa and Lisi, Giuseppe and Cortese, Aurelio and Watanabe, Takeo and Sasaki, Yuka and Kawato, Mitsuo},
  journal={Neuroimage},
  volume={188},
  pages={539--556},
  year={2019},
  publisher={Elsevier}
}

@article{cortese2022adaptive,
title = {Metacognitive resources for adaptive learning},
author={Cortese, Aurelio},
journal={Neuroscience Research},
year={2022},
volume={178}, 
pages={10-19},
doi={10.1016/j.neures.2021.09.003}

}

@article{petersAzimiTheory,
	title        = {How brains build higher order representations of uncertainty},
	author       = {Peters, Megan A. K. and Azimi Asrari, Hojjat},
	year         = 2026,
	journal = {Philosophy and the Mind Sciences},
    doi     = {10.33735/phimisci.2026.12269},
    url     ={https://philosophymindscience.org/index.php/phimisci/article/view/12269},
    volume  = {7},
    number  = {1},
}

@article{cortese2020unconscious,
	title        = {Unconscious reinforcement learning of hidden brain states supported by confidence},
	author       = {Cortese, Aurelio and Lau, Hakwan and Kawato, Mitsuo},
	year         = 2020,
	journal      = {Nature Communications},
	volume       = 11,
	number       = 1,
	pages        = 4429,
	doi          = {10.1038/s41467-020-17828-8},
	url          = {https://www.nature.com/articles/s41467-020-17828-8}
}

@article{Pedregosa2011ScikitLearn,
	title        = {Scikit-learn: Machine Learning in Python},
	author       = {Pedregosa, Fabian and Varoquaux, Ga{\"e}l and Gramfort, Alexandre and Michel, Vincent and Thirion, Bertrand and Grisel, Olivier and Blondel, Mathieu and Prettenhofer, Peter and Weiss, Ron and Dubourg, Vincent and VanderPlas, Jake and Passos, Alexandre and Cournapeau, David and Brucher, Matthieu and Perrot, Matthieu and Duchesnay, {\'E}douard},
	year         = 2011,
	journal      = {Journal of Machine Learning Research},
	volume       = 12,
	pages        = {2825--2830}
}

@inproceedings{shahdoust2024interictal,
	title        = {Interictal Epileptiform Discharges Disrupt Neural Computations Underlying Cognitive Control and Value-based Decision Making},
	author       = {Shahdoust, Niloufar and Cowan, Rhiannon L and Price, T Alexander and Kundu, Bornali and Davis, Tyler S and Rolston, John D and Rahimpour, SH and Smith, Elliot H},
	year         = 2024,
	booktitle    = {Proceedings of the Cognitive Computational Neuroscience Meeting}
}

@article{libowitz2025increased,
	title        = {Increased Aperiodic Exponents Track Depression Symptom Severity},
	author       = {Libowitz, Mark R and Sun, Wendy and Rabinovich, Rikki and Du, Jingnan and Campbell, Justin M and Cowan, Rhiannon L and Shahdoust, Niloufar and Price, T Alexander and Davis, Tyler S and Buckner, Randy L and others},
	year         = 2025,
	journal      = {bioRxiv},
	publisher    = {Cold Spring Harbor Laboratory},
	pages        = {2025--12}
}

@article{mathis2024decoding,
  title={Decoding the brain: From neural representations to mechanistic models},
  author={Mathis, Mackenzie Weygandt and Rotondo, Adriana Perez and Chang, Edward F and Tolias, Andreas S and Mathis, Alexander},
  journal={Cell},
  volume={187},
  number={21},
  pages={5814--5832},
  year={2024},
  publisher={Elsevier}
}

@article{song2020denoising,
  title={Denoising Diffusion Implicit Models},
  author={Song, Jiaming and Meng, Chenlin and Ermon, Stefano},
  journal={arXiv preprint arXiv:2010.02502},
  year={2020}
}

@article{rombach2022high,
  title={High-Resolution Image Synthesis with Latent Diffusion Models},
  author={Rombach, Robin and Blattmann, Andreas and Lorenz, Dominik and Esser, Patrick and Ommer, Bj{\"o}rn},
  journal={Proceedings of the IEEE/CVF Conference on Computer Vision and Pattern Recognition},
  pages={10684--10695},
  year={2022}
}

\clearpage
% \section*{Supplementary}
% \input{sections/supplementary}

\end{CJK}\end{document}